\documentclass[10pt,twocolumn,letterpaper]{article}

\usepackage[accsupp]{axessibility}  
\usepackage{iccv}
\usepackage{times}
\usepackage{epsfig}
\usepackage{graphicx}
\usepackage{amsmath}
\usepackage{amssymb}

\usepackage{tabularx}
\usepackage{multirow}
\usepackage[table,xcdraw,dvipsnames]{xcolor}
\usepackage[normalem]{ulem}
\useunder{\uline}{\ul}{}
\usepackage{mathtools}
\usepackage{bbm}

\usepackage{multirow}
\usepackage{diagbox}
\usepackage{booktabs}
\newcommand{\rb}{\rotatebox{90}}
\usepackage[caption=false]{subfig}
\renewcommand{\paragraph}[1]{\vspace{1mm}\noindent\textbf{#1}}
\usepackage{pdfpages}

\usepackage[pagebackref=true,breaklinks=true,letterpaper=true,colorlinks,bookmarks=false]{hyperref}

\iccvfinalcopy 

\def\iccvPaperID{4185} 

\ificcvfinal\fi

\begin{document}

\title{Hierarchical Memory Matching Network for Video Object Segmentation}

\author{Hongje Seong\textsuperscript{1} \quad\quad Seoung Wug Oh\textsuperscript{2} \quad\quad Joon-Young Lee\textsuperscript{2} \\ Seongwon Lee\textsuperscript{1} \quad\quad Suhyeon Lee\textsuperscript{1} \quad\quad Euntai Kim\textsuperscript{1,}\thanks{Corresponding author.}\vspace*{0.2cm}\\
{\textsuperscript{1}Yonsei University \quad\quad\quad \textsuperscript{2}Adobe Research}}

\maketitle
\ificcvfinal\thispagestyle{empty}\fi

\begin{abstract}
We present Hierarchical Memory Matching Network (HMMN) for semi-supervised video object segmentation. 
Based on a recent memory-based method~\cite{Oh_2019_ICCV}, we propose two advanced memory read modules that enable us to perform memory reading in multiple scales while exploiting temporal smoothness. We first propose a kernel guided memory matching module that replaces the non-local dense memory read, commonly adopted in previous memory-based methods. The module imposes the temporal smoothness constraint in the memory read, leading to accurate memory retrieval. More importantly, we introduce a hierarchical memory matching scheme and propose a top-k guided memory matching module in which memory read on a fine-scale is guided by that on a coarse-scale. With the module, we perform memory read in multiple scales efficiently and leverage both high-level semantic and low-level fine-grained memory features to predict detailed object masks.
Our network achieves state-of-the-art performance on the validation sets of DAVIS 2016/2017 (90.8\% and 84.7\%) and YouTube-VOS 2018/2019 (82.6\% and 82.5\%), and test-dev set of DAVIS 2017 (78.6\%).
The source code and model are available online: \url{https://github.com/Hongje/HMMN}.
\end{abstract}

\vspace{-5mm}
\section{Introduction}
\label{sec:1.Introduction}

\begin{figure}[t]
\centering
\includegraphics[width=1.\linewidth]{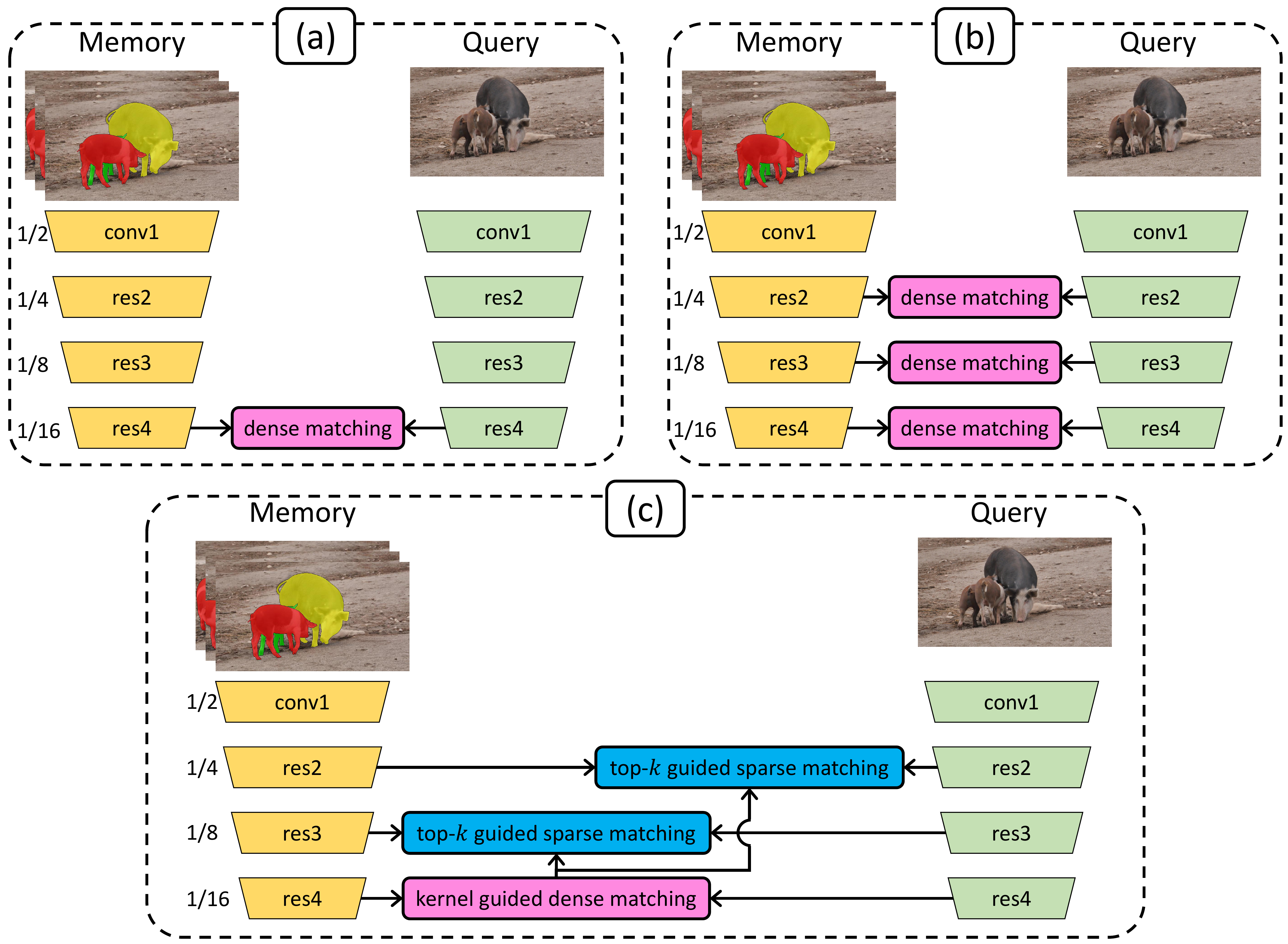}
\caption{
Previous memory-based methods densely match image features only at a coarse resolution, as shown in (a).
To conduct the memory reading at multiple scales, one can naively apply dense matching at each scale (b), but it needs prohibitive computational cost and is not robust due to noisy low-level features.
In our hierarchical memory matching architecture, shown in (c), fine-scale matching is guided by coarse-scale matching, resulting in efficient and robust memory matching in multiple scales.
\vspace{-0.5cm}
}
\label{fig:fig1}
\end{figure}

Semi-supervised video object segmentation (VOS) aims to predict the foreground object mask in every frame of a video given an object mask at the first frame.  Recently, memory-based VOS methods \cite{Oh_2019_ICCV,seong2020kernelized,lu2020video,li2020fast,li2020delving,liang2020video} have achieved great success. A key idea of the memory-based methods is matching densely between \textit{query} (\ie, current frame) and \textit{memory} (\ie, past frames with given or predicted masks) to retrieve the memory at a pixel-level.
Since the camera's field of view or objects in a video may move, spatio-temporal non-local and dense matching was performed to compute similarity for all matching possibilities.

There are two limitations of the existing memory-based methods: \textit{temporal smoothness} and \textit{fine-grained memory information}. Temporal smoothness is one of the strong constraints that we can assume for the VOS task. Previous VOS methods without memory often applied a local matching~\cite{voigtlaender2019feelvos,yang2020collaborative} or local refinement~\cite{perazzi2017learning,khoreva2019lucid,wug2018fast,yang2018efficient,hu2018videomatch,Zhang_2019_ICCV} between two adjacent frames for temporal smoothness. However, in the memory-based method~\cite{Oh_2019_ICCV}, the non-local matching completely ignores the constraint and it raises the risk of false matches (\eg, when multiple similar instances exist, see Fig.~\ref{fig:kernel_guidance_illustration}).
Another weakness is the lack of fine-grained memory information. In the memory-based methods, a query encoder only takes the current frame without any target information. Thus the memory matching is the only source to get information of the target object mask. The previous memory-based methods conduct the memory matching only at the coarsest resolution, (\eg, $1/16$ of the input resolution~\cite{Oh_2019_ICCV}), as shown in Fig. \ref{fig:fig1} (a). At the low resolution, while accurate matching is possible with high-level semantic features, we cannot expect fine-grained information that is also important to predict fine-detailed masks.

In this paper, we propose Hierarchical Memory Matching Network (HMMN) with two novel memory matching modules. To exploit the temporal smoothness, we propose \textit{kernel guided memory matching} module. We restrict possible correspondences between two adjacent frames to a local window and apply kernel guidance to the non-local memory matching that imposes the temporal smoothness constraint. For long-range matching between distant frames, we track the most probable correspondence for each memory pixel to a query pixel and apply relaxed kernel guidance according to the temporal distance, resulting in a smooth transition from local to global memory matching. This module replaces the non-local memory reading in the previous memory-based networks.

To retrieve fine-grained memory information, we propose \textit{top-k guided memory matching} module.
The computational cost for the dense memory matching grows quadratically with increasing search space.
Naively performing the memory reading at fine-scales \cite{yang2021collaborative} (Fig.~\ref{fig:fig1} (b)) requires prohibitively heavy computation. Also, memory matching with the low-level features at a fine-scale is susceptible to noisy matches.
Our top-$k$ guided memory matching solves both the computational cost and the matching robustness issues.
We first sample the top-$k$ candidate memory locations for each query pixel using the matching similarity score at the coarse-scale. Then, we conduct fine-scale memory matching between each query pixel and the corresponding candidate memory locations, as shown in Fig. \ref{fig:fig1} (c). The top-$k$ guided memory matching reduces the matching complexity at high-resolution significantly from $\mathcal{O}(TH^{2}W^{2})$ to $\mathcal{O}(kHW)$, where $T$, $H$, and $W$ are the time, height, and width of the feature map, and $k$ is a constant. The coarse-to-fine hierarchical matching scheme makes our fine-scale memory matching robust even with low-level features.
We note that some previous works \cite{lai2020mast,zhu2021deformable} also reduce memory matching complexity by extracting $k$ matching candidates but they select candidates using features at the same scale.
In contrast, we selected $k$ matching candidates from high-level (\ie, coarse-scale) semantic features, thus semantically more accurate matching candidates would be selected.

Our contributions are summarized as follows:
\vspace{-\topsep}
\begin{itemize}
\setlength\itemsep{-0.3em}
\item We propose kernel guided memory matching module, imposing the temporal smoothness constraint to the non-local matching with all memory frames.
\item We propose top-$k$ guided memory matching module, resulting in efficient and robust fine-scale memory matching.
\item With the two novel memory matching modules, we present Hierarchical Memory Matching Network (HMMN) that performs coarse-to-fine hierarchical memory matching effectively.
\item Our network achieves state-of-the-art performance on both DAVIS and YouTube-VOS benchmarks.
\end{itemize}

\section{Related Work}
\label{sec:2.Related_Work}
\vspace{-3mm}
\paragraph{Semi-supervised Video Object Segmentation:}
\label{sec:2.1.Semi_supervised_Video_Object_Segmentation}
Semi-supervised VOS \cite{perazzi2016benchmark,pont20172017,xu2018youtube} has been tackled in two ways: online-learning method and offline-learning method.
The online-learning methods \cite{cheng2017segflow,caelles2017one,voigtlaender2017online,bao2018cnn,luiten2018premvos,maninis2018video,8611188,Duarte_2019_ICCV,meinhardt2020make} fine-tune networks at test time using the given ground-truth mask at the first frame.
The objective of fine-tuning is to let networks detect target objects for each video.
Therefore, the online-learning method can expect accurate results by training a target-specific network, but they are subject to severe disadvantages at run-time because the network needs to be trained multiple times on the first frame during testing.

Offline-learning methods aim to train a network that works well for any input videos without test-time training.
It has usually been solved by mask propagation or pixel-wise matching.
The propagation-based methods~\cite{perazzi2017learning,khoreva2019lucid,hu2017maskrnn,li2018video,wug2018fast,johnander2019generative,chen2020state,zhang2020transductive,hu2020dipnet} train a network to propagate the given mask sequentially from the first frame.
Since the propagation is conducted in a short-time interval, the methods often exploit the temporal smoothness constraint but are not robust to occlusion.
The matching-based methods~\cite{shin2017pixel,hu2018videomatch,Zeng_2019_ICCV,voigtlaender2019feelvos,huang2020fast,yang2020collaborative} predict a foreground mask in the current frame based on matching with previously predicted or given mask.
Recently, STM~\cite{Oh_2019_ICCV} introduced a memory-based method for offline-learning VOS and demonstrated a significantly improved performance while achieving a fast run-time.
Our approach follows the memory-based method, and we address the main limitations of existing methods.

\begin{figure*}[t]
\centering
\includegraphics[width=0.86\linewidth]{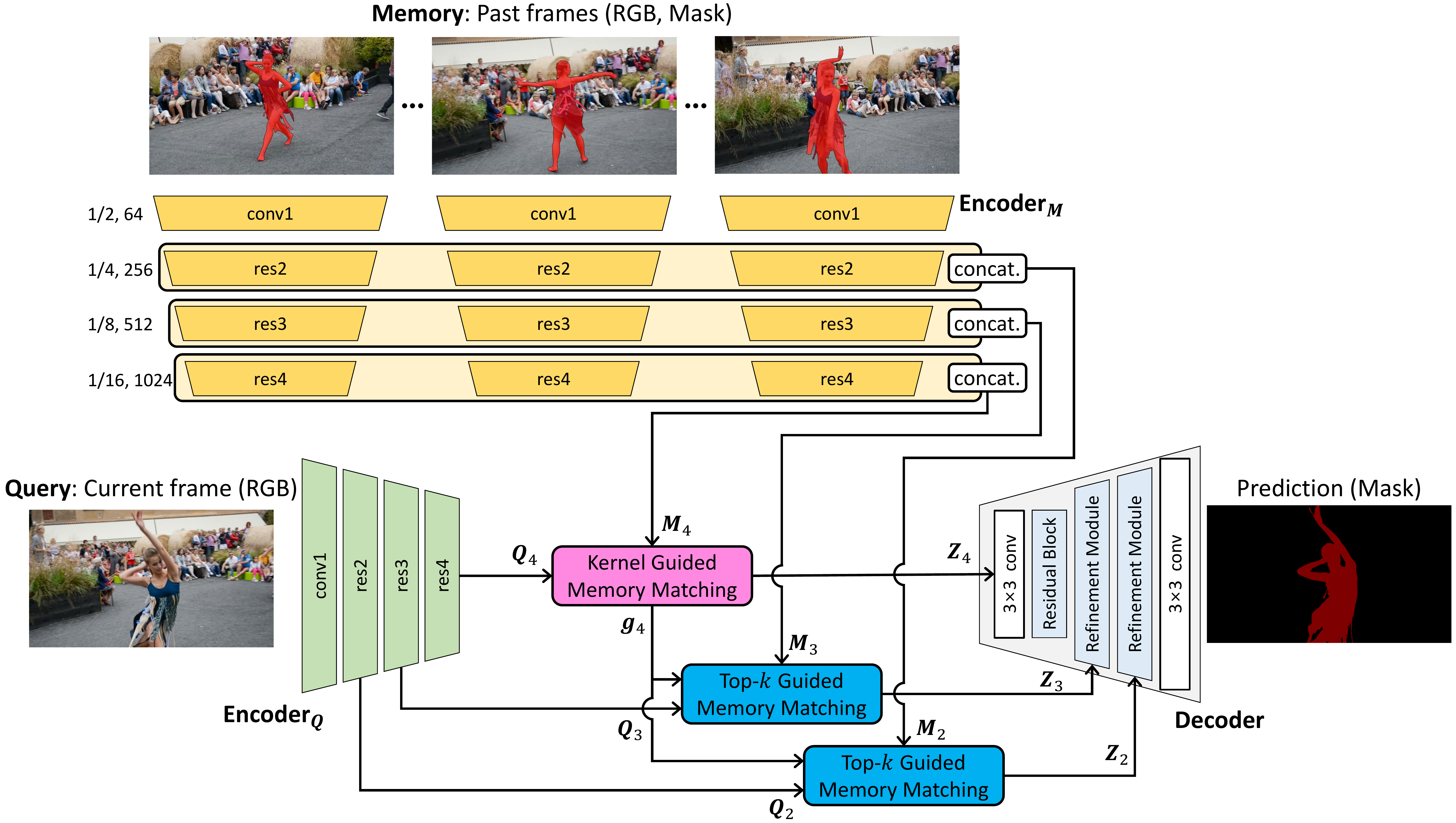}
\caption{
An overview of HMMN. Our network consists of two ResNet-based encoders for the query and the memory frames that extract multi-scale features, kernel guided memory matching block that operates on the coarsest scale, top-$k$ guided memory matching blocks that operate on the finer scales, and a decoder that takes the memory reading results and produces the final mask prediction.
\vspace{-0.5cm}
}
\label{fig:overall_architecture}
\end{figure*}

\paragraph{Memory-based Video Object Segmentation:}
\label{sec:2.2.Memory_based_Video_Object_Segmentation}
Memory networks \cite{sukhbaatar2015end,miller2016key,kumar2016ask} memorize external information as \textbf{key} and \textbf{value}, then the \textbf{value} is retrieved by query via non-local matching with the \textbf{key}.
It was first proposed for natural language processing, and STM \cite{Oh_2019_ICCV} repurposed the memory networks to memory-based VOS.
STM retrieves memory using non-local and dense memory matching and finds the target object in the query using the retrieved memory.
Extended from STM, EGMN~\cite{lu2020video} proposed the graph memory networks to update memory using query.
GC~\cite{li2020fast} introduced a new global matching method for fast memory matching.
Liang~\etal~\cite{liang2020video} proposed an adaptive memory update scheme to reduce redundant computation at memory matching.
Li~\etal~\cite{li2020delving} explored a cyclic mechanism for both training and inference to boost performance.
KMN~\cite{seong2020kernelized} additionally conducted \textit{memory-to-query} matching then applied 2D Gaussian kernels on the query for robust matching.
The previous memory-based methods overlook the temporal smoothness, one of the most important cues for VOS, as they performed memory matching in a non-local manner. 
In addition, the previous works conduct memory matching only at the coarsest resolution, which hard to expect to take fine mask information.
We address the problems by introducing two matching modules, kernel guided memory matching and top-$k$ memory matching.
Note that our kernel guided memory matching is completely different from kernelization used in KMN \cite{seong2020kernelized}, which generates kernel based on non-local matching thus does not exploit temporal smoothness.

\section{Method}
\label{sec:3.HMMN}
Our method, Hierarchical Memory Matching Network (HMMN), is based on STM \cite{Oh_2019_ICCV}. 
Given a ground-truth object mask at the first frame, we sequentially predict the target object mask from the second frame to the last frame.
The past frames concatenated with predicted or given masks are set to memory, and the current frame is used as a query.

The main distinction comes from the construction and the use of hierarchical memory.
The objective of the hierarchical memory is to leverage memories in multiple scales, from low-resolution semantic features to high-resolution detailed features, on the memory-based VOS architectures.
To efficiently read the information from the hierarchical memory, we design two types of memory matching modules based on the feature map's scale: kernel guided dense memory matching at the coarsest scale, and top-$k$ guided sparse memory matching at fine scales.
At the coarsest scale, we perform dense and non-local query-memory matching similar to STM~\cite{Oh_2019_ICCV} and other variants.
But, we improve the robustness of the global matching through the kernel guidance that exploits temporal smoothness as an additional cue.
At the finer scales followed by the coarsest level, we perform a sparse query-memory matching making use of the matching results from the coarsest level as guidance. 
Specifically, we take the top-$k$ memory matching for each query point at the coarsest scale and use them to guide the sparse matching at the finer scales.
In this way, we can retrieve fine-detailed memory information while taking a fractional computational cost compared to dense memory matching. 

The overview of our network is shown in Fig.~\ref{fig:overall_architecture}.
In our network, memory and query frames are first fed into two independent ResNet50~\cite{b23}-based encoders.
Both encoders extract multi-scale features -- $\mathbf{Q}_S$ for the query frame and $\mathbf{M}_S$ for the memory frames -- from ResNet50's $S$-th \texttt{res} block. 
We use three scales where $S \in \{2,3,4\}$ with the output scale of \{1/4, 1/8, 1/16\} with respect to the input image.
At each scale, in the order of coarse-to-fine scales, we perform a memory read by matching the query and memory features, and then the outputs further go through the decoder to predict an object mask.

For the memory matching in the coarsest scale, the embedded query and memory \{$\mathbf{Q}_4$, $\mathbf{M}_4$\} are fed into \textit{kernel guided memory matching} module, and it outputs the updated feature ($\mathbf{Z}_4$) and a guidance ($\mathbf{g}_4$) which is the similarity matrix used for memory retrieval.
For the finer scales ($S$ is 2 or 3), \textit{top-$k$ guided memory matching} module is used instead.
It takes a pair of embedded query and memory \{$\mathbf{Q}_S$, $\mathbf{M}_S$\} along with the guidance ($\mathbf{g}_4$), and outputs the updated feature $\mathbf{Z}_S$.
Finally, the decoder takes all the output features $\mathbf{Z}_S$ (either as the input or through a skip-connection), and makes a mask prediction.
Note that, except for the new memory matching modules, we keep the rest of the network structure (\eg, encoder and decoder design) as the same as STM~\cite{Oh_2019_ICCV}.

\begin{figure}[t]
\centering
\includegraphics[width=1.\linewidth]{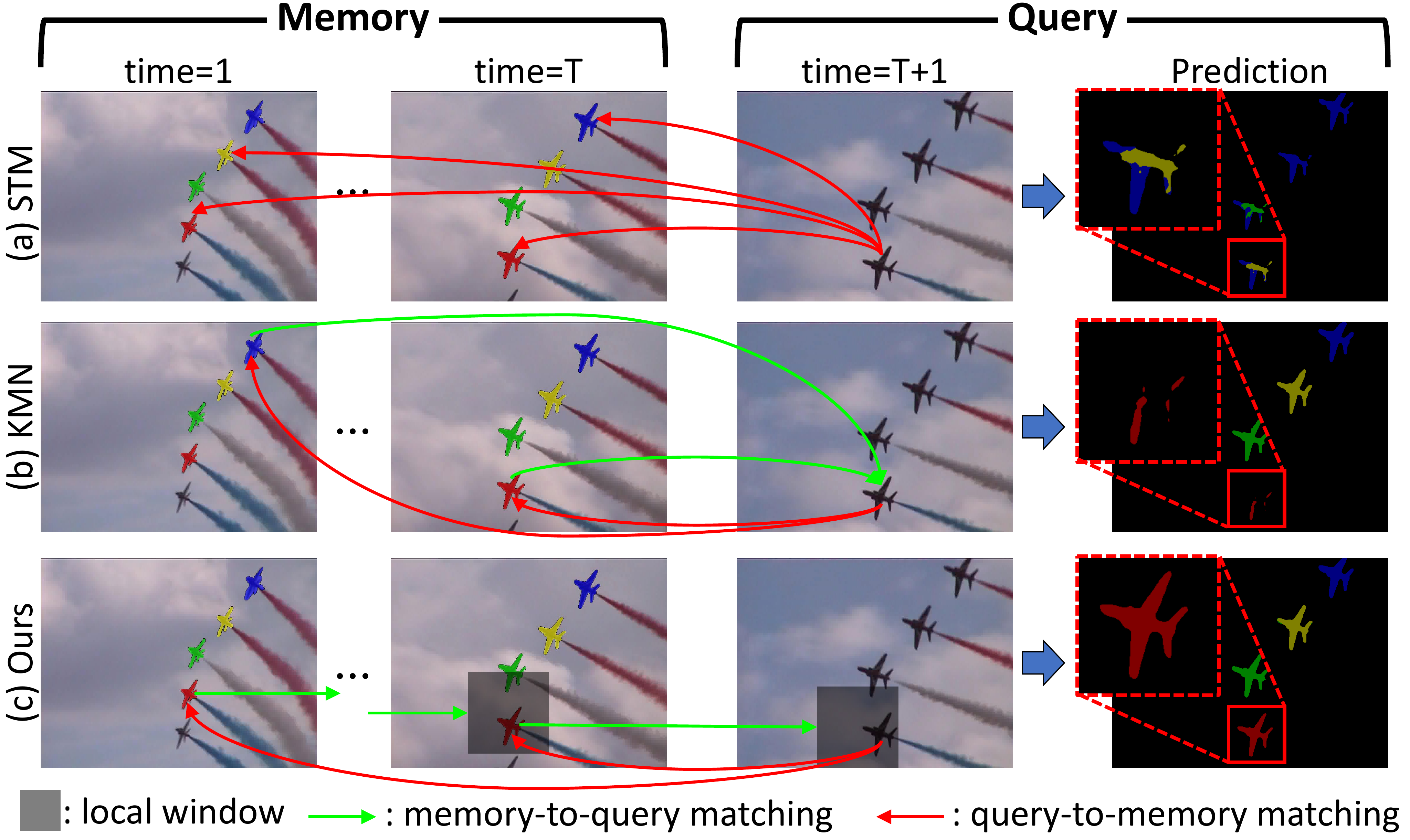}
\caption{
The effect of our kernel guided matching module. (a) STM does not make use of bi-directional memory-to-query matching. (b) KMN performs a memory-to-query matching in a non-local manner, thus it cannot make use of temporal smoothness prior. (c) Our memory-to-query matching is achieved by connecting local tracking, thus it can impose temporal smoothness on the query-to-memory matching results. 
\vspace{-0.5cm}
}
\label{fig:kernel_guidance_illustration}
\end{figure}

\begin{figure}[t]
\centering
\includegraphics[width=1.\linewidth]{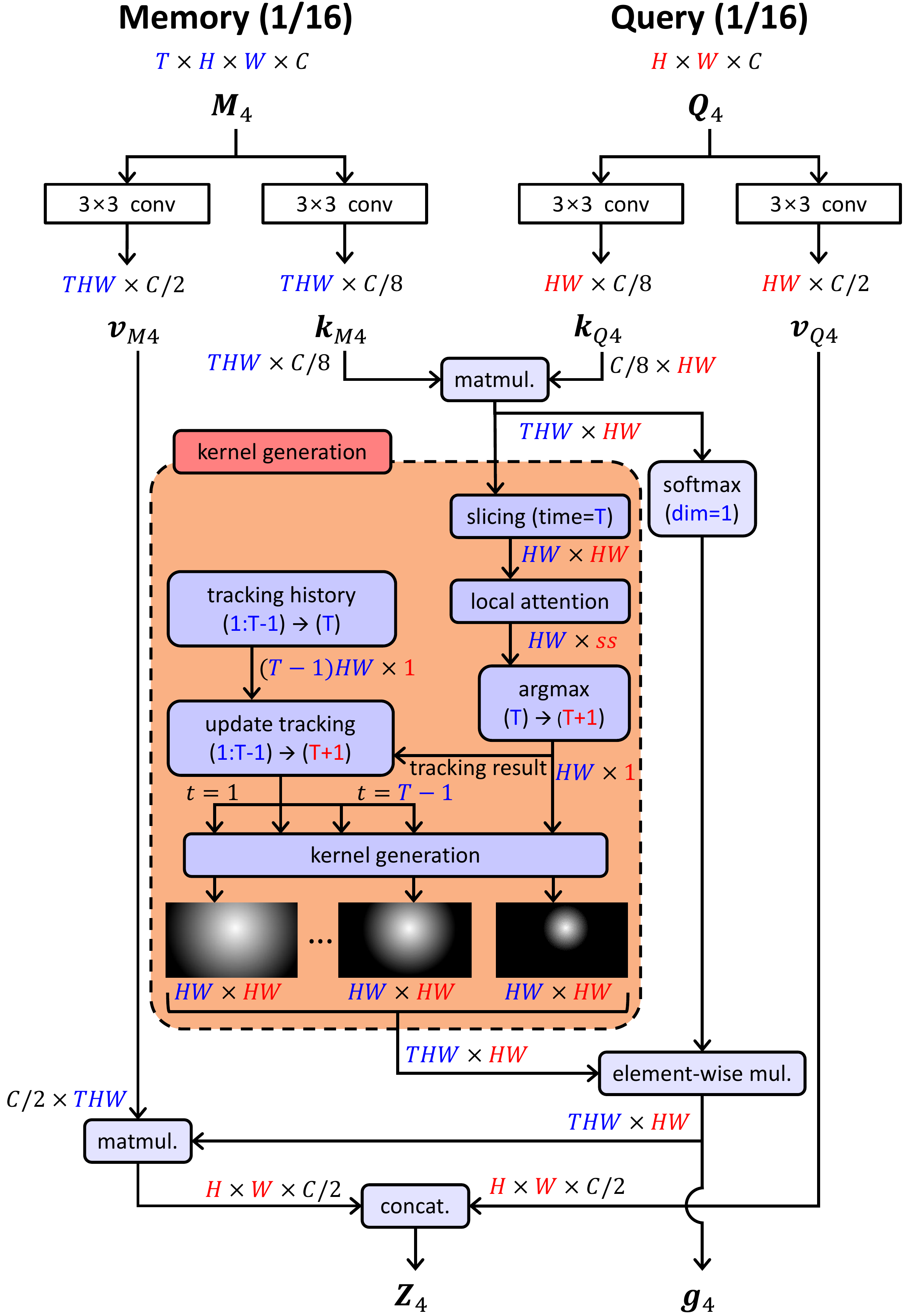}
\caption{
A detailed implementation of kernel guided memory matching module.
We use \textcolor{blue}{blue} and \textcolor{red}{red} to indicate memory and query dimensions, respectively. 
Note that we can access the tracking history (1:T-1)$\rightarrow$(T) that is saved beforehand, thus only the tracking between the previous frame and the current frame (T)$\rightarrow$ (T+1) is needed to be newly computed. 
}
\vspace{-0.5cm}
\label{fig:kernel_guidance_memory_read_detail}
\end{figure}

\subsection{Kernel Guided Memory Matching}
\label{sec:3.1.Kernel_Guided_Memory_Matching}
With the embedded memory and query ($\mathbf{M}_4, \mathbf{Q}_4$), extracted from each encoder at \texttt{res4} stage, we first encode \textbf{keys} ($\mathbf{k}_{M4}$, $\mathbf{k}_{Q4}$) and \textbf{values} ($\mathbf{v}_{M4}$, $\mathbf{v}_{Q4}$) via four independent $3 \times 3$ convolutional layers.
Then, a non-local matching between memory and query is performed using \textbf{keys} as follows:
\begin{equation}
\mathcal{M}_{4} = \mathbf{k}_{M4} \mathbf{k}_{Q4}^\top ,
\label{eq1}
\end{equation}
where $^\top$ indicates a matrix transpose. 
Based on the non-local matching ($\mathcal{M}_{4}$), we compute the attention map ($\mathbf{g}_{4}$) by
\begin{equation}
\mathbf{g}_{4} = {L_1}\left( {\mathcal{K}(\mathcal{M}_{4}) \odot \text{softmax}({\mathcal{M}}_4)} \right),
\label{eq2}
\end{equation}
where $\odot$ indicates an element-wise multiplication, $L_1( \cdot )$ is L1 normalization which normalizes along the memory dimension, and $\mathcal{K}( \cdot )$ is 2D Gaussian kernel.
Then, the memory \textbf{value} is retrieved using the attention map ($\mathbf{g}_{4}$) as follows:
\begin{equation}
\mathbf{v}'_{M4} = \mathbf{v}_{M4}^\top \mathbf{g}_{4} .
\label{eq3}
\end{equation}
Finally, the query \textbf{value} ($\mathbf{v}_{Q4}$) is concatenated with the retrieved \textbf{value} ($\mathbf{v}'_{M4}$) along the feature dimension to be the output.

Here, we impose the temporal smoothness, that is the common and strong constraint for videos, on the memory matching through the kernel prior ($\mathcal{K}$).
If $\mathcal{K}( \cdot ) = \mathbf{1}$, the output ($\mathbf{Z}_4$) will be the same as the output from vanilla memory read block used in STM \cite{Oh_2019_ICCV}.
In other words, STM \cite{Oh_2019_ICCV} retrieves memory solely based on non-local query-to-memory matching (\ie, $\text{softmax} (\mathcal{M}_{4})$), as illustrated in Fig.~\ref{fig:kernel_guidance_illustration}~(a).
Thus, the fact that objects are likely to appear in similar local positions between adjacent frames (\ie, temporal smoothness) is completely ignored.
To harness this behavior, we additionally generate a kernel guidance ($\mathcal{K}( \cdot )$) based on spatio-temporal local matching.
As illustrated in Fig.~\ref{fig:kernel_guidance_illustration}~(c), we conduct memory-to-query matching between two adjacent frames for every memory pixel.
Here, we constrained the matching to perform only within a local region with a window size of $s$.
Between every two adjacent frames, we track every pixel by selecting a single pixel within a local window that has the highest similarity score.
This way, every memory pixel can reach to the best-matching query pixel by connecting local pixel-level tracking frame-by-frame.
Based on the resulting memory-to-query matching, we generate 2D Gaussian kernels for every memory pixel with the standard deviation of $\sigma^t$.
As the temporal distance of memory-to-query increases, the tracking error can be accumulated and the temporal smoothness constraint weakens. Thus, we relaxed the kernel guidance by controlling the standard deviation according to the temporal distance by ${\sigma ^t} = {\sigma _{init}} + (T - t){\sigma _{factor}}$.
This results in a smooth transition from local to global memory matching according to the temporal distance between query and memory features.
A detailed implementation of kernel guided memory matching module is shown in Fig.~\ref{fig:kernel_guidance_memory_read_detail}.

Note that our kernel guidance is inspired by KMN \cite{seong2020kernelized}, but the objective is completely different.
KMN~\cite{seong2020kernelized} used kernel only for robust matching from bi-directional attention, thus the kernels were generated based on non-local matching, as illustrated in Fig.~\ref{fig:kernel_guidance_illustration}~(b).
Our kernel guidance, however, is based on fully local matching, and it effectively exploits the temporal smoothness as shown in Fig.~\ref{fig:kernel_guidance_illustration}~(c).

\subsection{Top-$k$ Guided Memory Matching}
\label{sec:3.3.Topk_guided_Memory_Read}
The main objective of computing a dense spatio-temporal attention map in memory matching module is to find when-and-where each query pixel attends to memory pixel.
However, computing the dense attention map in high resolution requires prohibitively large computing resources as its computational complexity grows quadratically with regard to the feature map size.
Thus, computing dense attention maps for finer levels of the feature hierarchy (\texttt{res3} and \texttt{res2}) is computationally too expensive.
We address this issue by reducing the number of matching candidates in memory using top-$k$ guidance.

\begin{figure}[t]
\centering
\includegraphics[width=1.\linewidth]{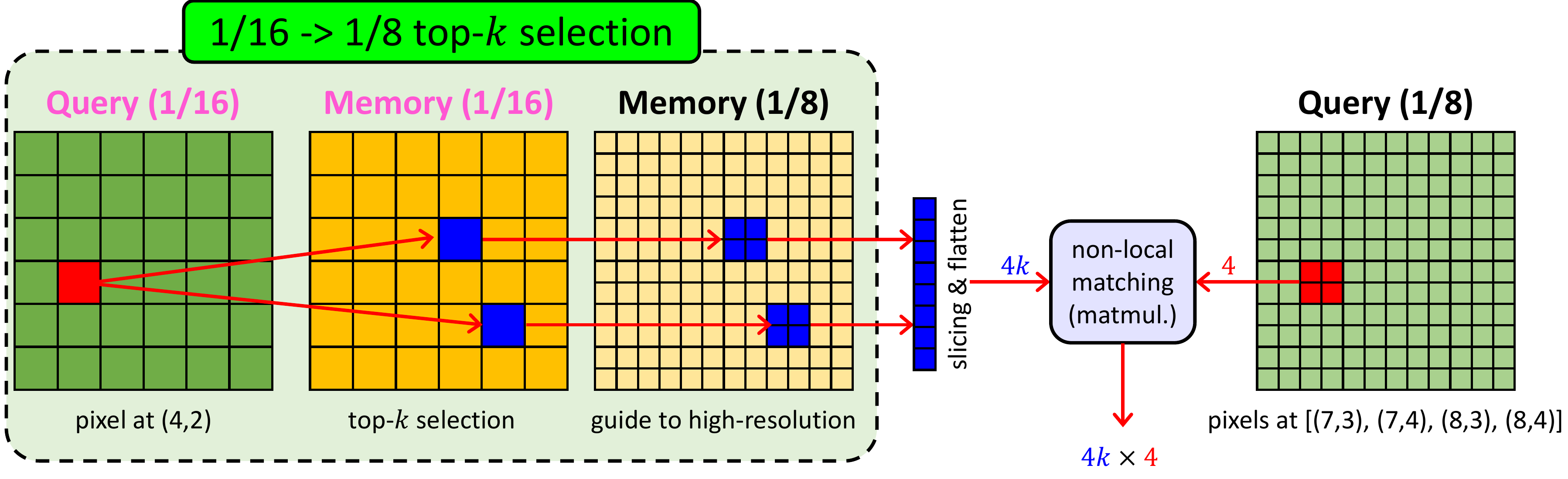}
\caption{
An example of top-$k$ selection at \texttt{res3} stage. To simplify the illustration, $k$ is set to 2 in this example. 
}
\vspace{-0.5cm}
\label{fig:topk_4k_selection}
\end{figure}

\begin{figure}[t]
\centering
\includegraphics[width=1.\linewidth]{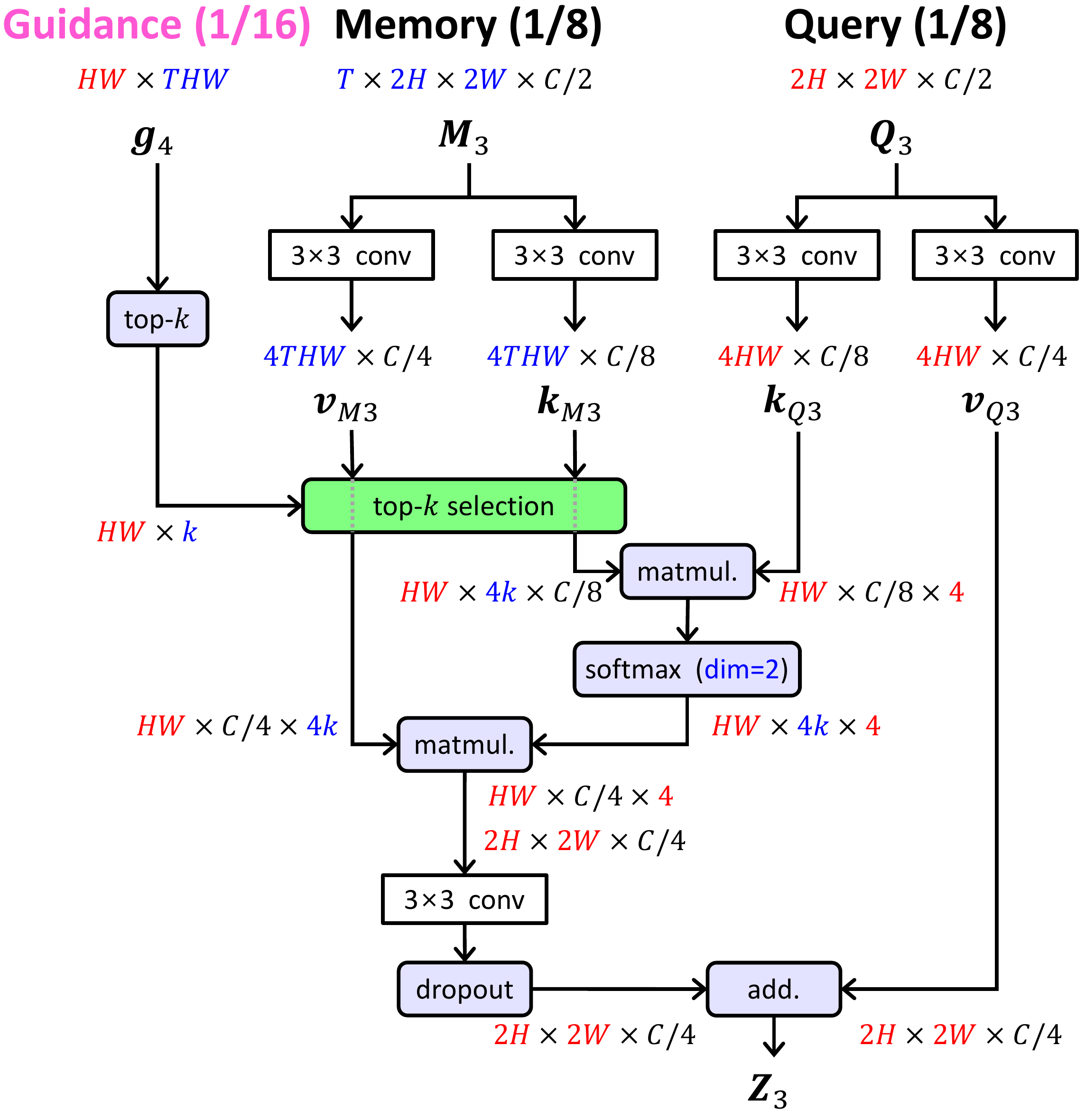}
\caption{
A detailed implementation of the top-$k$ guided memory matching module at the \texttt{res3} stage.
Memory and query dimensions are indicated using \textcolor{blue}{blue} and \textcolor{red}{red}.
The detailed implementation at the \texttt{res2} stage is provided in the supplementary material.
}
\vspace{-0.5cm}
\label{fig:topk_memory_read_detail}
\end{figure}

Here, we assume that the matching result at high-resolution should be similar to that at low-resolution.
By this assumption, we reuse the dense matching result at low-resolution as guidance for matching in higher resolution.
An illustration of selecting $k$ pixels and guiding to high-resolution for each query pixel is depicted in Fig. \ref{fig:topk_4k_selection}.
Based on the low-resolution attention map ($\mathbf{g}_{4}$), which comes from \texttt{res4} stage, we select $k$ best matching memory pixels for each query pixel via top-$k$ operation.
Then, only a sparse matching to selected pixels from the memory is performed.

Note that the selected $k$ pixels in \texttt{res4} correspond to 4$k$ and 16$k$ pixels at \texttt{res3} and \texttt{res2} stages, respectively, thus we take $k$ and $k/4$ for guiding each at \texttt{res3} and \texttt{res2} stage in order to have a similar computational overhead.
This memory read module based on sparse matching can be efficiently implemented with a combination of common tensor operations.
A detailed implementation of the top-$k$ guided memory matching module is shown in Fig.~\ref{fig:topk_memory_read_detail}.
The outputs of top-$k$ guided memory matching modules ($\textbf{Z}_3$, $\textbf{Z}_2$) are fed into the decoder through shortcut connections at the corresponding scale.

Note that, in the module, rather than directly using the retrieved \textbf{values} as the output, we place one convolutional layer followed by Dropout layer before added to the query \textbf{value} as residual. 
This design choice is due to the following observation. Without dropout, the model tends to converge to a sub-optimal state that does not make use of the matching results at the coarsest scale (\ie, memory at \texttt{res4}).
This sub-optimal model appears to take a shortcut for easier solutions, simply relying on low-level mask information (\ie, memory at \texttt{res2} and \texttt{res3}) ignoring the high-level semantic matching.
We were able to prevent this behavior by delivering the information in a restrictive way through a residual connection after a dropout layer that randomly drops the \textit{whole} input feature during training.
In this way, the network has to consider the output of top-$k$ guided memory matching module as supplementary information to refine the memory matching at the coarsest resolution.

\section{Experiments}
\label{sec:4.Experiments}
\subsection{Implementation Details}
\label{sec:4.1.Implementation_Details}
\paragraph{Training.}~For a fair comparison with STM~\cite{Oh_2019_ICCV}, we follow the same training strategies.
We initialize the encoders with ImageNet~\cite{b19} pre-trained weights and randomly initialize the other layers.
Then, we take the images with object masks in \cite{b42,b43,hariharan2011semantic,shi2015hierarchical,cheng2014global,wang2017salient} and pretrain HMMN on the image datasets.
Specifically, we generate three frames by augmenting each image via random affine transforms.
The random affine transforms include rotation, shearing, zooming, translation, and cropping.
During the pre-training, the dropout rate in top-$k$ guided memory matching module (\S\ref{sec:3.3.Topk_guided_Memory_Read}) is gradually decreased from 1 to 0.5.

After the pre-training on image datasets, the main training is done using either DAVIS 2017~\cite{pont20172017} or YouTube-VOS 2019~\cite{xu2018youtube} training set depending on the target benchmark.
During main training, three frames are randomly sampled from a video with the gradually increasing maximum interval (from 0 to 25).
The dropout rate in top-$k$ guided memory matching module is gradually decreased from 0.5 to 0.

During both pre-training and main training, we minimize pixel-wise cross-entropy loss with Adam optimizer \cite{KingmaB14}, and the learning rate is set to 1e-5.
We use an input size of $384 \times 384$ and a mini-batch size of 4.
According to \cite{Oh_2019_ICCV}, we employ the soft aggregation operation when multiple target objects exist in a video.

\paragraph{Inference.}~As in \cite{Oh_2019_ICCV,seong2020kernelized}, we take the first frame, the previous frame, and the intermediate frames sampled at every 5 frames for the memory in the coarsest scale ($\mathbf{M}_4$). 
For the fine-scale memories ($\mathbf{M}_3, \mathbf{M}_2$), we do not use the intermediate frames to avoid GPU memory overflow unless mentioned otherwise.
We use the same number of $k$ for top-$k$ guided memory matching during training and inference, which is set to 32.
The kernel guidance in \S\ref{sec:3.1.Kernel_Guided_Memory_Matching} is used only during inference, as in KMN \cite{seong2020kernelized}.
We have tried to use the kernel guidance during training, but there was no noticeable improvement.
We set the standard deviation of $\sigma_{init}$ and $\sigma_{factor}$ into 3 and 0.5, respectively, and we used window size $s$ of 7.
We measure our run-time using a single NVIDIA GeForce 1080 Ti GPU.

\begin{table}[t]
\begin{center}
\footnotesize

\begin{tabular}{lccccc}
\toprule
\multicolumn{1}{c}{Method}                              & OL         & $\mathcal{J\&F}$ & $\mathcal{J}$ & $\mathcal{F}$ & Time     \\
\midrule
e-OSVOS   \cite{meinhardt2020make}                      & \checkmark & 86.8             & 86.6          & 87.0          & 3.4$s$   \\
DyeNet \cite{li2018video}                               & \checkmark & -                & 86.2          & -             & 2.32$s$  \\
RaNet \cite{Wang_2019_ICCV}                             & \checkmark & 87.1             & 86.6          & 87.6          & 4$s$     \\
STM \textbf{(+YV)}   \cite{Oh_2019_ICCV}                &            & 89.3             & 88.7          & 89.9          & 0.16$s$  \\
CFBI \textbf{(+YV)}   \cite{yang2020collaborative}      &            & 89.4             & 88.3          & 90.5          & 0.18$s$  \\
KMN \textbf{(+YV)}   \cite{seong2020kernelized}         &            & 90.5             & 89.5          & 91.5          & 0.12$s$  \\
\midrule
HMMN                                                    &            & 89.4             & 88.2          & 90.6          & 0.10$s$  \\
HMMN \textbf{(+YV)}                                     &            & \textbf{90.8}    & \textbf{89.6} & \textbf{92.0} & 0.10$s$  \\
\bottomrule
\end{tabular}
\vspace{-0.2cm}
\end{center}
\caption{Comparison on DAVIS 2016 validation set.
(\textbf{+YV}) indicates YouTube-VOS is additionally used for training, and OL denotes the use of online-learning strategies during test-time. Time measurements reported in this table are directly from the corresponding papers.}
\vspace{-0.5cm}
\label{tab:davis2016_val}
\end{table}

\begin{table}[t]
\begin{center}
\footnotesize

\begin{tabular}{lcccc}
\toprule
\multicolumn{1}{c}{Method}                              & OL         & $\mathcal{J\&F}$      & $\mathcal{J}$         & $\mathcal{F}$         \\
\midrule
FRTM \textbf{(+YV)}   \cite{robinson2020learning}       & \checkmark           & 76.7                  & -                     & -                     \\
e-OSVOS   \cite{meinhardt2020make}                      & \checkmark & 77.2                  & 74.4                  & 80.0                  \\
PReMVOS   \cite{luiten2018premvos}                      & \checkmark & 77.8                  & 73.9                  & 81.7                  \\
LWL \textbf{(+YV)}   \cite{bhat2020learning}            &            & 81.6                  & 79.1                  & 84.1                  \\
STM \textbf{(+YV)}   \cite{Oh_2019_ICCV}                &            & 81.8                  & 79.2                  & 84.3                  \\
CFBI \textbf{(+YV)}   \cite{yang2020collaborative}      &            & 81.9                  & 79.1                  & 84.6                  \\
EGMN \textbf{(+YV)}   \cite{lu2020video}                &            & 82.8                  & 80.2                  & 85.2                  \\
KMN \textbf{(+YV)}   \cite{seong2020kernelized}         &            & 82.8                  & 80.0                  & 85.6                  \\
\midrule
HMMN                                                    &            &80.4 & 77.7 & 83.1
 \\
HMMN \textbf{(+YV)}                                     &            & \textbf{84.7}         & \textbf{81.9}         & \textbf{87.5}        \\
\bottomrule
\end{tabular}
\end{center}
\vspace{-0.2cm}
\caption{Comparison on DAVIS 2017 validation set.}
\vspace{-0.2cm}
\label{tab:davis2017_val}
\end{table}

\begin{table}[t]
\begin{center}
\footnotesize

\begin{tabular}{lcccc}
\toprule
\multicolumn{1}{c}{Method}                              & OL         & $\mathcal{J\&F}$ & $\mathcal{J}$ & $\mathcal{F}$ \\
\midrule
CINN \cite{bao2018cnn}                                  & \checkmark & 67.5             & 64.5          & 70.5          \\
DyeNet \cite{li2018video}                               & \checkmark & 68.2             & 65.8          & 70.5          \\
PReMVOS   \cite{luiten2018premvos}                      & \checkmark & 71.6             & 67.5          & 75.7          \\
STM \textbf{(+YV)}   \cite{Oh_2019_ICCV}                &            & 72.2             & 69.3          & 75.2          \\
CFBI \textbf{(+YV)}   \cite{yang2020collaborative}      &            & 74.8             & 71.1          & 78.5          \\
KMN \textbf{(+YV)}   \cite{seong2020kernelized}         &            & 77.2             & 74.1          & 80.3          \\
\midrule
HMMN \textbf{(+YV)}                                     &            & \textbf{78.6}    & \textbf{74.7} & \textbf{82.5}\\
\bottomrule
\end{tabular}
\end{center}
\vspace{-0.2cm}
\caption{Comparison on DAVIS 2017 test-dev set.}
\vspace{-0.5cm}
\label{tab:davis2017_test_dev}
\end{table}

\subsection{Comparisons}
\label{sec:4.2.Comparisons}
We compare our HMMN against state-of-the-art methods on DAVIS~\cite{perazzi2016benchmark,pont20172017} and YouTube-VOS~\cite{xu2018youtube} benchmarks.
For DAVIS benchmarks, 60 videos from DAVIS 2017 training set are used during main training following the common evaluation protocol~\cite{wug2018fast,yang2018efficient,Oh_2019_ICCV,seong2020kernelized}.
In addition, we report our results on DAVIS benchmarks using additional training videos from Youtube-VOS for a fair comparison with some recent methods~\cite{Oh_2019_ICCV,seong2020kernelized,bhat2020learning,yang2020collaborative,lu2020video,robinson2020learning}.
For Youtube-VOS benchmarks, the training set of 3471 videos are used. 
For all experiments, we either use the official evaluation code or upload our results to the evaluation server.

\paragraph{DAVIS}~\cite{perazzi2016benchmark,pont20172017}
is a densely annotated VOS dataset and mostly adopted benchmark to evaluate VOS models.
To evaluate HMMN on DAVIS benchmarks, we use an input size of 480p resolution for all experiments.
DAVIS dataset is divided into two sets: (1) DAVIS 2016, which is an object-level annotated dataset (single object); and (2) DAVIS 2017, which is an instance-level annotated dataset (multiple objects).
The official metrics, region similarity $\mathcal{J}$ and contour accuracy $\mathcal{F}$, are measured for comparison.
As shown in Table~\ref{tab:davis2016_val}, our HMMN achieves state-of-the-art performance while taking a fast run-time on DAVIS 2016 validation set.
Further, even without an additional YouTube-VOS dataset to train HMMN, we surpass most state-of-the-art methods.

We also conduct comparisons on DAVIS 2017 validation and test-dev sets, and the results are given in Table~\ref{tab:davis2017_val} and Table~\ref{tab:davis2017_test_dev}.
As shown in the Tables, our HMMN significantly outperforms the current best results by \textbf{1.9}\% and \textbf{1.4}\% of $\mathcal{J \& F}$ scores on DAVIS 2017 validation and test-dev sets, respectively.
We omitted some comparable works in the tables. The full comparison tables are available in the supplementary material.

\begin{table}[t]
\begin{center}
\footnotesize

\begin{tabular}{lcccccc}
\toprule
\multicolumn{1}{c}{Method}             & OL         & $\mathcal{G}$ & $\mathcal{J_S}$ & $\mathcal{J_U}$ & $\mathcal{F_S}$ & $\mathcal{F_U}$ \\
\midrule
& & \multicolumn{5}{c}{YouTube-VOS 2018   validation set}                                                                                       \\
\midrule
AGSS-VOS \cite{Lin_2019_ICCV}          &            & 71.3          & 71.3            & 65.5            & 75.2            & 73.1            \\
e-OSVOS   \cite{meinhardt2020make}     & \checkmark & 71.4          & 71.7            & 74.3            & 66.0            & 73.8            \\
FRTM   \cite{robinson2020learning}     & \checkmark           & 72.1          & 72.3            & 65.9            & 76.2            & 74.1            \\
STG-Net   \cite{liu2020spatiotemporal} &            & 73.0          & 72.7            & 69.1            & 75.2            & 74.9            \\
STM \cite{Oh_2019_ICCV}                &            & 79.4          & 79.7            & 72.8            & 84.2            & 80.9            \\
AFB+URR \cite{liang2020video}          &            & 79.6          & 78.8            & 74.1            & 83.1            & 82.6            \\
EGMN \cite{lu2020video}                &            & 80.2          & 80.7            & 74.0            & 85.1            & 80.9            \\
CFBI   \cite{yang2020collaborative}    &            & 81.4          & 81.1            & 75.3            & 85.8            & 83.4            \\
KMN \cite{seong2020kernelized}         &            & 81.4          & 81.4            & 75.3            & 85.6            & 83.3            \\
LWL \cite{bhat2020learning}            &            & 81.5          & 80.4            & 76.4            & 84.9            & 84.4            \\
\midrule
HMMN                                   &            & \textbf{82.6} & \textbf{82.1}   & \textbf{76.8}   & \textbf{87.0}   & \textbf{84.6}   \\
\bottomrule
& & \multicolumn{5}{c}{YouTube-VOS 2019   validation set}                                                                                       \\
\midrule
STM* \cite{Oh_2019_ICCV}               &            & 79.3          & 79.8            & 73.0            & 83.8            & 80.5            \\
\multicolumn{2}{l}{KMN*   \cite{seong2020kernelized}}                 & 80.0          & 80.4            & 73.8            & 84.5            & 81.4            \\
CFBI   \cite{yang2020collaborative}    &            & 81.0          & 80.6            & 75.2            & 85.1            & 83.0            \\
\midrule
HMMN                                   &            & \textbf{82.5} & \textbf{81.7}   & \textbf{77.3}   & \textbf{86.1}   & \textbf{85.0}  \\
\bottomrule
\end{tabular}
\end{center}
\vspace{-0.2cm}
\caption{Comparison on YouTube-VOS validation sets.
$\mathcal{G}$ is an average of $\mathcal{J_S}$, $\mathcal{J_U}$, $\mathcal{F_S}$, and $\mathcal{F_U}$.
* denotes our reproduced result using our training setup.
\vspace{-0.5cm}
}
\label{tab:youtube_vos_val}
\end{table}

\paragraph{YouTube-VOS}~\cite{xu2018youtube}
is a large-scale benchmark for VOS.
To evaluate our HMMN on YouTube-VOS benchmarks, we reduce the input image to 480p resolution.
We measured region similarity ($\mathcal{J_S}, \mathcal{J_U}$) and contour accuracy ($\mathcal{F_U}, \mathcal{F_U}$) for 65 of seen and 26 of unseen object categories separately.
In Table~\ref{tab:youtube_vos_val}, we compare HMMN with state-of-the-art methods on YouTube-VOS 2018 and 2019 validation sets.
Note that only CFBI \cite{yang2020collaborative} officially reported for comparison on YouTube-VOS 2019 validation set, so we additionally report our reproduced results of STM~\cite{Oh_2019_ICCV} and KMN~\cite{seong2020kernelized} using our training setup.
As shown in Table~\ref{tab:youtube_vos_val}, our HMMN surpasses the state-of-the-art methods in all official metrics on both YouTube-VOS 2018 and 2019.

\begin{figure*}[t]
\centering
\includegraphics[width=1\linewidth]{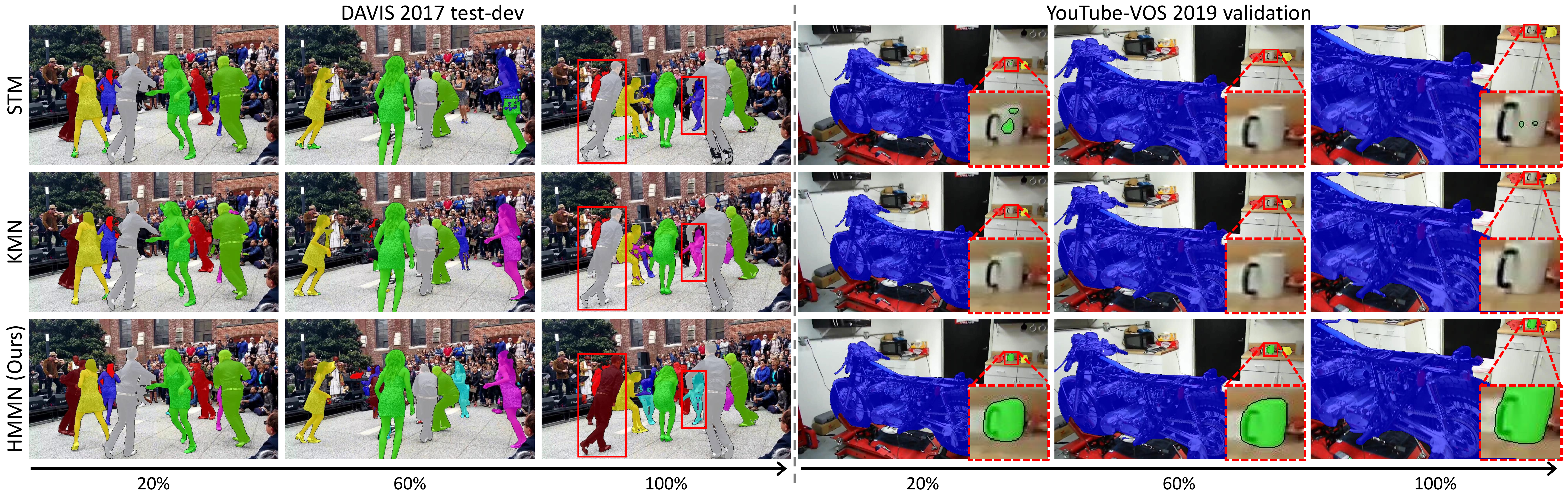}
\caption{Qualitative comparison on DAVIS 2017 test-dev, and Youtube-VOS 2019 validation sets.
We compare HMMN with STM \cite{Oh_2019_ICCV} and KMN \cite{seong2020kernelized}.
We marked significant improvements from STM and KMN using red boxes.
}
\label{fig:qualitative_results}
\end{figure*}

\paragraph{Qualitative Comparison.}~Fig.~\ref{fig:qualitative_results} shows qualitative comparison with STM \cite{Oh_2019_ICCV} and KMN \cite{seong2020kernelized}.
In the figure, STM \cite{Oh_2019_ICCV} almost failed to predict target objects when multiple similar objects have appeared or several occlusion occurred (DAVIS example).
KMN~\cite{seong2020kernelized} failed to predict a very small object (YouTube-VOS example).
On the other hand, our HMMN predicted the target objects accurately in the challenging cases.
More qualitative results are provided in the supplementary material.

\begin{table}[t]
\begin{center}
\footnotesize

\begin{tabular}{ccc|ccc|cc}
\toprule
                           &            &            & \multicolumn{3}{c|}{DAVIS}              & \multicolumn{2}{c}{YouTube-VOS} \\
K*~\cite{seong2020kernelized} & \textbf{K} & \textbf{T} & Time    & 2016          & 2017          & 2018           & 2019           \\
\midrule
                           &            &            & 0.07$s$ & 89.2          & 82.2          & 79.2           & 79.3           \\
\checkmark                 &            &            & 0.07$s$ & 89.5          & 83.3          & 79.8           & 80.0           \\
                           & \checkmark &            & 0.07$s$ & 90.0          & 83.1          & 80.7           & 80.9           \\
                           &            & \checkmark & 0.10$s$ & \textbf{90.8} & 83.6          & 81.1           & 81.2           \\
\checkmark                 &            & \checkmark & 0.10$s$ & \textbf{90.8} & 84.1          & 81.7           & 81.8           \\
                           & \checkmark & \checkmark & 0.10$s$ & \textbf{90.8} & \textbf{84.7} & \textbf{82.6}  & \textbf{82.5}\\
                           \bottomrule
\end{tabular}
\end{center}
\vspace{-0.2cm}
\caption{\textbf{Module ablation study}.
We report $\mathcal{J \& F}$ and $\mathcal{G}$ scores for DAVIS and Youtube-VOS, respectively.
The run-time is measured on DAVIS 2016 validation set.
The baseline model is STM \cite{Oh_2019_ICCV}.
K*~\cite{seong2020kernelized} denotes the kernelization proposed in \cite{seong2020kernelized}, and \textbf{K} and \textbf{T} indicate our kernel guided memory matching and top-$k$ guided memory matching modules, respectively.
\vspace{-0.5cm}
}
\label{tab:ablation_study}
\end{table}

\begin{table*}[t]
\begin{center}
\footnotesize
\begin{tabular}{cccc}
\multirow{2}{*}{
\subfloat[Comparison on various $k$-pixel selection strategies.
\label{tab:ablation_k_pixel_selection}]
{\begin{tabular}{cc|c}
& \vspace{-1cm} & \\
\toprule
                              & $k$             &                         \\
                              \midrule
                              & 0              & 83.1/80.9 \\
                              \midrule
\multirow{3}{*}{\rb{random}}  & 64              & 82.9/80.5                   \\
                              & 128             & 82.2/80.3                   \\
                              & 256             & 81.9/81.1                   \\
                              \midrule
\multirow{2}{*}{\rb{stride}}  & 2$\cdot$8$^2$  & 82.4/80.3                   \\
                              & 2$\cdot$12$^2$ & 83.3/80.5                   \\
                              \midrule
\multirow{3}{*}{\rb{top-$k$}} & 8               & 82.2/81.4                   \\
                              & 16              & 83.2/82.0                   \\
                              & 32              & \textbf{84.7}/\textbf{82.4}\\
                              \bottomrule
\end{tabular}
} } &
\hspace{0.01\linewidth} 
\subfloat[Results with large number of $k$.
\label{tab:ablation_topk_vs_dense}]
{\begin{tabular}{lc|cc}
\toprule 
                                &          & \multicolumn{2}{c}{Training}            \\
                                & $k$      & 32                          & $\infty$  \\
                                \midrule
\multirow{5}{*}{\rb{Inference}} & 32       & 84.7/\textbf{82.5}          & 82.3/81.5 \\
                                & 64       & \textbf{84.9}/\textbf{82.5} & 82.6/81.5 \\
                                & 128      & 84.6/82.1                   & 82.8/81.4 \\
                                & 256      & 84.3/81.9                   & 82.7/81.5 \\
                                & $\infty$ & 82.0/78.8                   & 82.1/80.5\\
                                \bottomrule 
\end{tabular}
}  &
\hspace{0.01\linewidth}
\subfloat[Experimental results with and without dropout in top-$k$ memory matching module.
\label{tab:ablation_pyramid_stages}]
{\begin{tabular}{lc}
& \vspace{-1.75cm} \\
\toprule
Training    &                         \\
\midrule
w/o Dropout & 81.5/80.7                  \\
w/ Dropout  & \textbf{84.7}/\textbf{82.5} \\
\bottomrule
\end{tabular}
} &
\hspace{0.01\linewidth}
\subfloat[Comparison of memory management strategies for top-$k$ memory matching module.
\label{tab:memory_interval}]
{\begin{tabular}{lc}
& \vspace{-1.75cm} \\
\toprule
\multicolumn{2}{l}{Fine-scale memory frames}                         \\
\midrule
First   \& Prev. & 84.7/\textbf{82.5}          \\
Every 5 frames   & \textbf{84.9}/\textbf{82.5}\\
\bottomrule
\end{tabular}
} \\[-1mm] &
\hspace{0.01\linewidth}
\subfloat[Memory stages for top-$k$ memory matching module.
\label{tab:ablation_dropout}]
{\begin{tabular}{lc}
\toprule
\multicolumn{2}{l}{Fine-scale memory stages}                         \\
\midrule
None                             & 83.1/80.9                   \\
\texttt{res2}                    & 83.2/82.1                   \\
\texttt{res3}                    & 83.2/81.6                   \\
\texttt{res2} \&   \texttt{3} & \textbf{84.7}/\textbf{82.5}\\
\bottomrule
\end{tabular}
} &
\hspace{0.01\linewidth}
\subfloat[Window sizes in kernel guided memory matching module.
\label{tab:ablation_window_size}]
{\begin{tabular}{lc}
& \vspace{-1cm} \\
\toprule
\multicolumn{2}{l}{Window   size ($s$)} \\
\midrule
3$\times$3      & 84.0/\textbf{82.5}    \\
5$\times$5      & 84.4/\textbf{82.5}    \\
7$\times$7      & 84.7/\textbf{82.5}    \\
9$\times$9      & \textbf{84.8}/82.4    \\
11$\times$11    & 84.7/82.3             \\
$\infty$        & 84.2/81.0            \\
\bottomrule
\end{tabular}
}  &
\subfloat[Standard deviations of Gaussian kernel in kernel guided memory matching module.
\label{tab:ablation_standard_deviation}]
{ \makebox[0.2\linewidth][c]{\begin{tabular}{cc}
& \vspace{-1cm} \\
\toprule
\multicolumn{2}{l}{Standard   deviation ($\sigma_{init}$)} \\
\midrule
1               & 83.6/82.0                                \\
3               & \textbf{84.7}/\textbf{82.5}              \\
5               & 84.0/82.4                                \\
7               & 83.8/82.4                                \\
9               & 83.8/82.2                                \\
11              & 83.7/81.9                               \\
\bottomrule
\end{tabular}
} }
\end{tabular}
\end{center}
\vspace{-0.2cm}
\caption{\textbf{Ablation Study.} For each setting, we report results of $\mathcal{J \& F}$ and $\mathcal{G}$ scores on DAVIS 2017 and YouTube-VOS 2019 validation sets, respectively.
\label{tab:ablation}
\vspace{-0.5cm}
}
\end{table*}

\subsection{Ablation Experiments}
\label{sec:4.4.Ablation_Experiments}
\paragraph{Module ablation.}~We conduct an ablation study on our two proposed memory matching modules to demonstrate the efficacy of those.
We also compare our kernel guided memory matching with the kernelization method proposed in KMN~\cite{seong2020kernelized}. 
As shown in Table \ref{tab:ablation_study}, our kernel guidance is more effective than one from KMN, and the use of fine-scale memories through top-$k$ guided memory module greatly boosts the performance to the state-of-the-art.

\paragraph{Temporal stability ($\mathcal{T}$).}~To validate the effectiveness of our HMMN on temporal smoothness quantitatively, we evaluate temporal stability ($\mathcal{T}$) \cite{perazzi2016benchmark} on DAVIS 2016 validation set.
STM~\cite{Oh_2019_ICCV}, KMN~\cite{seong2020kernelized}, and our HMMN achieved $\mathcal{T}$ scores (lower is better) of 17.2\%, 15.2\%, and 13.0\%, respectively.
This implies that our method significantly improves temporal stability over STM and KMN.

\paragraph{$k$-pixel selection strategies.}~To validate the effectiveness of our top-$k$ guidance (\S\ref{sec:3.3.Topk_guided_Memory_Read}), we study various strategies to sample $k$ memory pixels.
As can be seen in Table \ref{tab:ablation} (a), fine-scale memory with simple sampling methods (random, stride) do not provide consistent improvement over the baseline ($k$=0). However, fine-scale memory with our top-$k$ guidance yields significant performance improvement even with a small number of $k$.

\paragraph{The effect of $k$.}~We further study the effect of $k$ during both training and inference
by increasing the number of $k$ from 32 to $\infty$. 
Here, $k$=$\infty$ indicates using \textit{dense} memory without sampling.
As shown in Table \ref{tab:ablation} (b), using a dense fine-scale memory either in training and/or inference degrades the overall performance compared to using top-$k$ sampled memory. 
We conjecture that, in fine-scale, the feature is not robust enough for the global and dense matching. 
In this case, top-$k$ guidance could be beneficial to rejecting noises by restricting the search space into few reliable options. 
While our default setting is to set $k$=32 for both training and inference, we observed that the performance could be further improved by tuning $k$.

\paragraph{Dropout for high-resolution memory.}~Table \ref{tab:ablation} (c) shows the effect of dropout in top-$k$ guided memory module. 
As we discussed in \S \ref{sec:3.3.Topk_guided_Memory_Read}, our dropout strategy makes our network learn with hierarchical memories effectively.

\paragraph{Fine-scale memory management.}~Table~\ref{tab:ablation} (d) shows that we can further boost our performance by exploiting fine-scale memories from the intermediate frames sampled from every 5 frames.
However, this configuration requires too much GPU memory to store memory features, while performance improvement is marginal.
We use the first and previous frames for fine-scale memory by default to run HMMN. 
Note that we use the intermediate frames for the coarse-scale memory. 

\paragraph{Fine-scale memory stages.}~We ablate hierarchical memory stage-by-stage, and the results are given in Table \ref{tab:ablation} (e).
As shown in the table, using memory hierarchies in both stages shows the best performance.
If the hierarchical memory is used only in a single stage, interestingly, taking finer-scale memory (\ie, \texttt{res2} stage) achieves better performance even we reduced the number of $k$ to $k/4$ at \texttt{res2} stage. 
It is thought that the finer-scale memory can provide more complementary information to the memory from the coarsest scale.

\paragraph{Window size $s$ \& standard deviation $\sigma_{init}$.}~Tables \ref{tab:ablation} (f) and \ref{tab:ablation} (g) show the parameter search experiments for guidance kernel (\S\ref{sec:3.1.Kernel_Guided_Memory_Matching}).
Choosing a too large and too small value for $s$ and $\sigma_{init}$ has degraded the performance.
Therefore, we select proper window size $s$ and standard deviation of Gaussian kernel $\sigma_{init}$ as 7$\times$7 and 3, respectively.

\section{Conclusion}
\label{sec:5.Conclusion}
We presented two advanced memory matching modules that exploit temporal smoothness and hierarchical memory effectively.
We demonstrated the efficacy of our HMMN through extensive experiments and achieved state-of-the-art performance on all evaluated benchmarks while keeping a fast run-time.
We believe that our proposed two memory matching modules can be further extended to other matching-based vision applications such as video saliency detection, video instance segmentation, and semantic correspondence.

\paragraph{Acknowledgement.}~This work was supported by the Industry Core Technology Development Project, 20005062, Development of Artificial Intelligence Robot Autonomous Navigation Technology for Agile Movement in Crowded Space, funded by the Ministry of Trade, industry \& Energy (MOTIE, Republic of Korea).

{\small
\bibliographystyle{ieee_fullname}
\bibliography{main}

\begin{thebibliography}{10}\itemsep=-1pt

\bibitem{bao2018cnn}
Linchao Bao, Baoyuan Wu, and Wei Liu.
\newblock Cnn in mrf: Video object segmentation via inference in a cnn-based
  higher-order spatio-temporal mrf.
\newblock In {\em CVPR}, pages 5977--5986, 2018.

\bibitem{bhat2020learning}
Goutam Bhat, Felix~J{\"a}remo Lawin, Martin Danelljan, Andreas Robinson,
  Michael Felsberg, Luc Van~Gool, and Radu Timofte.
\newblock Learning what to learn for video object segmentation.
\newblock In {\em ECCV}, 2020.

\bibitem{caelles2017one}
Sergi Caelles, Kevis-Kokitsi Maninis, Jordi Pont-Tuset, Laura Leal-Taix{\'e},
  Daniel Cremers, and Luc Van~Gool.
\newblock One-shot video object segmentation.
\newblock In {\em CVPR}, pages 221--230, 2017.

\bibitem{chen2020state}
Xi Chen, Zuoxin Li, Ye Yuan, Gang Yu, Jianxin Shen, and Donglian Qi.
\newblock State-aware tracker for real-time video object segmentation.
\newblock In {\em CVPR}, pages 9384--9393, 2020.

\bibitem{cheng2017segflow}
Jingchun Cheng, Yi-Hsuan Tsai, Shengjin Wang, and Ming-Hsuan Yang.
\newblock Segflow: Joint learning for video object segmentation and optical
  flow.
\newblock In {\em ICCV}, pages 686--695, 2017.

\bibitem{cheng2014global}
Ming-Ming Cheng, Niloy~J Mitra, Xiaolei Huang, Philip~HS Torr, and Shi-Min Hu.
\newblock Global contrast based salient region detection.
\newblock {\em IEEE Transactions on Pattern Analysis and Machine Intelligence},
  37(3):569--582, 2014.

\bibitem{Duarte_2019_ICCV}
Kevin Duarte, Yogesh~S. Rawat, and Mubarak Shah.
\newblock Capsulevos: Semi-supervised video object segmentation using capsule
  routing.
\newblock In {\em ICCV}, October 2019.

\bibitem{b42}
Mark Everingham, Luc Van~Gool, Christopher~KI Williams, John Winn, and Andrew
  Zisserman.
\newblock The pascal visual object classes (voc) challenge.
\newblock {\em International Journal of Computer Vision}, 88(2):303--338, 2010.

\bibitem{hariharan2011semantic}
Bharath Hariharan, Pablo Arbel{\'a}ez, Lubomir Bourdev, Subhransu Maji, and
  Jitendra Malik.
\newblock Semantic contours from inverse detectors.
\newblock In {\em ICCV}, pages 991--998. IEEE, 2011.

\bibitem{b23}
Kaiming He, Xiangyu Zhang, Shaoqing Ren, and Jian Sun.
\newblock Deep residual learning for image recognition.
\newblock In {\em CVPR}, pages 770--778, 2016.

\bibitem{hu2020dipnet}
Ping Hu, Jun Liu, Gang Wang, Vitaly Ablavsky, Kate Saenko, and Stan Sclaroff.
\newblock Dipnet: Dynamic identity propagation network for video object
  segmentation.
\newblock In {\em WACV}, pages 1904--1913, 2020.

\bibitem{hu2017maskrnn}
Yuan-Ting Hu, Jia-Bin Huang, and Alexander Schwing.
\newblock Maskrnn: Instance level video object segmentation.
\newblock In {\em NIPS}, pages 325--334, 2017.

\bibitem{hu2018videomatch}
Yuan-Ting Hu, Jia-Bin Huang, and Alexander~G Schwing.
\newblock Videomatch: Matching based video object segmentation.
\newblock In {\em ECCV}, pages 54--70, 2018.

\bibitem{huang2020fast}
Xuhua Huang, Jiarui Xu, Yu-Wing Tai, and Chi-Keung Tang.
\newblock Fast video object segmentation with temporal aggregation network and
  dynamic template matching.
\newblock In {\em CVPR}, pages 8879--8889, 2020.

\bibitem{johnander2019generative}
Joakim Johnander, Martin Danelljan, Emil Brissman, Fahad~Shahbaz Khan, and
  Michael Felsberg.
\newblock A generative appearance model for end-to-end video object
  segmentation.
\newblock In {\em CVPR}, pages 8953--8962, 2019.

\bibitem{khoreva2019lucid}
Anna Khoreva, Rodrigo Benenson, Eddy Ilg, Thomas Brox, and Bernt Schiele.
\newblock Lucid data dreaming for video object segmentation.
\newblock {\em International Journal of Computer Vision}, 127(9):1175--1197,
  2019.

\bibitem{KingmaB14}
Diederik~P. Kingma and Jimmy Ba.
\newblock Adam: {A} method for stochastic optimization.
\newblock In {\em ICLR}, 2015.

\bibitem{kumar2016ask}
Ankit Kumar, Ozan Irsoy, Peter Ondruska, Mohit Iyyer, James Bradbury, Ishaan
  Gulrajani, Victor Zhong, Romain Paulus, and Richard Socher.
\newblock Ask me anything: Dynamic memory networks for natural language
  processing.
\newblock In {\em ICML}, pages 1378--1387, 2016.

\bibitem{lai2020mast}
Zihang Lai, Erika Lu, and Weidi Xie.
\newblock Mast: A memory-augmented self-supervised tracker.
\newblock In {\em CVPR}, pages 6479--6488, 2020.

\bibitem{li2018video}
Xiaoxiao Li and Chen Change~Loy.
\newblock Video object segmentation with joint re-identification and
  attention-aware mask propagation.
\newblock In {\em ECCV}, pages 90--105, 2018.

\bibitem{li2020fast}
Yu Li, Zhuoran Shen, and Ying Shan.
\newblock Fast video object segmentation using the global context module.
\newblock In {\em ECCV}, 2020.

\bibitem{li2020delving}
Yuxi Li, Ning Xu, Jinlong Peng, John See, and Weiyao Lin.
\newblock Delving into the cyclic mechanism in semi-supervised video object
  segmentation.
\newblock In {\em NIPS}, 2020.

\bibitem{liang2020video}
Yongqing Liang, Xin Li, Navid Jafari, and Qin Chen.
\newblock Video object segmentation with adaptive feature bank and
  uncertain-region refinement.
\newblock In {\em NIPS}, 2020.

\bibitem{Lin_2019_ICCV}
Huaijia Lin, Xiaojuan Qi, and Jiaya Jia.
\newblock Agss-vos: Attention guided single-shot video object segmentation.
\newblock In {\em ICCV}, October 2019.

\bibitem{b43}
Tsung-Yi Lin, Michael Maire, Serge Belongie, James Hays, Pietro Perona, Deva
  Ramanan, Piotr Doll{\'a}r, and C~Lawrence Zitnick.
\newblock Microsoft coco: Common objects in context.
\newblock In {\em ECCV}, pages 740--755. Springer, 2014.

\bibitem{liu2020spatiotemporal}
Daizong Liu, Shuangjie Xu, Xiao-Yang Liu, Zichuan Xu, Wei Wei, and Pan Zhou.
\newblock Spatiotemporal graph neural network based mask reconstruction for
  video object segmentation.
\newblock In {\em AAAI}, 2021.

\bibitem{lu2020video}
Xinkai Lu, Wenguan Wang, Martin Danelljan, Tianfei Zhou, Jianbing Shen, and Luc
  Van~Gool.
\newblock Video object segmentation with episodic graph memory networks.
\newblock In {\em ECCV}, 2020.

\bibitem{luiten2018premvos}
Jonathon Luiten, Paul Voigtlaender, and Bastian Leibe.
\newblock Premvos: Proposal-generation, refinement and merging for video object
  segmentation.
\newblock In {\em ACCV}, pages 565--580. Springer, 2018.

\bibitem{maninis2018video}
K-K Maninis, Sergi Caelles, Yuhua Chen, Jordi Pont-Tuset, Laura Leal-Taix{\'e},
  Daniel Cremers, and Luc Van~Gool.
\newblock Video object segmentation without temporal information.
\newblock {\em IEEE Transactions on Pattern Analysis and Machine Intelligence},
  41(6):1515--1530, 2019.

\bibitem{meinhardt2020make}
Tim Meinhardt and Laura Leal-Taix{\'e}.
\newblock Make one-shot video object segmentation efficient again.
\newblock In {\em NIPS}, 2020.

\bibitem{miller2016key}
Alexander Miller, Adam Fisch, Jesse Dodge, Amir-Hossein Karimi, Antoine Bordes,
  and Jason Weston.
\newblock Key-value memory networks for directly reading documents.
\newblock In {\em EMNLP}, 2016.

\bibitem{wug2018fast}
Seoung~Wug Oh, Joon-Young Lee, Kalyan Sunkavalli, and Seon~Joo Kim.
\newblock Fast video object segmentation by reference-guided mask propagation.
\newblock In {\em CVPR}, pages 7376--7385, 2018.

\bibitem{Oh_2019_ICCV}
Seoung~Wug Oh, Joon-Young Lee, Ning Xu, and Seon~Joo Kim.
\newblock Video object segmentation using space-time memory networks.
\newblock In {\em ICCV}, October 2019.

\bibitem{perazzi2017learning}
Federico Perazzi, Anna Khoreva, Rodrigo Benenson, Bernt Schiele, and Alexander
  Sorkine-Hornung.
\newblock Learning video object segmentation from static images.
\newblock In {\em CVPR}, pages 2663--2672, 2017.

\bibitem{perazzi2016benchmark}
Federico Perazzi, Jordi Pont-Tuset, Brian McWilliams, Luc Van~Gool, Markus
  Gross, and Alexander Sorkine-Hornung.
\newblock A benchmark dataset and evaluation methodology for video object
  segmentation.
\newblock In {\em CVPR}, pages 724--732, 2016.

\bibitem{pont20172017}
Jordi Pont-Tuset, Federico Perazzi, Sergi Caelles, Pablo Arbel{\'a}ez, Alex
  Sorkine-Hornung, and Luc Van~Gool.
\newblock The 2017 davis challenge on video object segmentation.
\newblock {\em arXiv preprint arXiv:1704.00675}, 2017.

\bibitem{robinson2020learning}
Andreas Robinson, Felix~Jaremo Lawin, Martin Danelljan, Fahad~Shahbaz Khan, and
  Michael Felsberg.
\newblock Learning fast and robust target models for video object segmentation.
\newblock In {\em CVPR}, pages 7406--7415, 2020.

\bibitem{b19}
Olga Russakovsky, Jia Deng, Hao Su, Jonathan Krause, Sanjeev Satheesh, Sean Ma,
  Zhiheng Huang, Andrej Karpathy, Aditya Khosla, Michael Bernstein, et~al.
\newblock Imagenet large scale visual recognition challenge.
\newblock {\em International Journal of Computer Vision}, 115(3):211--252,
  2015.

\bibitem{seong2020kernelized}
Hongje Seong, Junhyuk Hyun, and Euntai Kim.
\newblock Kernelized memory network for video object segmentation.
\newblock In {\em ECCV}, 2020.

\bibitem{shi2015hierarchical}
Jianping Shi, Qiong Yan, Li Xu, and Jiaya Jia.
\newblock Hierarchical image saliency detection on extended cssd.
\newblock {\em IEEE Transactions on Pattern Analysis and Machine Intelligence},
  38(4):717--729, 2015.

\bibitem{shin2017pixel}
Jae Shin~Yoon, Francois Rameau, Junsik Kim, Seokju Lee, Seunghak Shin, and In
  So~Kweon.
\newblock Pixel-level matching for video object segmentation using
  convolutional neural networks.
\newblock In {\em CVPR}, pages 2167--2176, 2017.

\bibitem{sukhbaatar2015end}
Sainbayar Sukhbaatar, Jason Weston, Rob Fergus, et~al.
\newblock End-to-end memory networks.
\newblock In {\em NIPS}, pages 2440--2448, 2015.

\bibitem{voigtlaender2019feelvos}
Paul Voigtlaender, Yuning Chai, Florian Schroff, Hartwig Adam, Bastian Leibe,
  and Liang-Chieh Chen.
\newblock Feelvos: Fast end-to-end embedding learning for video object
  segmentation.
\newblock In {\em CVPR}, pages 9481--9490, 2019.

\bibitem{voigtlaender2017online}
Paul Voigtlaender and Bastian Leibe.
\newblock Online adaptation of convolutional neural networks for video object
  segmentation.
\newblock In {\em BMVC}, 2017.

\bibitem{wang2017salient}
Jingdong Wang, Huaizu Jiang, Zejian Yuan, Ming-Ming Cheng, Xiaowei Hu, and
  Nanning Zheng.
\newblock Salient object detection: A discriminative regional feature
  integration approach.
\newblock {\em International Journal of Computer Vision}, 123(2):251--268,
  2017.

\bibitem{Wang_2019_ICCV}
Ziqin Wang, Jun Xu, Li Liu, Fan Zhu, and Ling Shao.
\newblock Ranet: Ranking attention network for fast video object segmentation.
\newblock In {\em ICCV}, October 2019.

\bibitem{8611188}
Huaxin Xiao, Bingyi Kang, Yu Liu, Maojun Zhang, and Jiashi Feng.
\newblock Online meta adaptation for fast video object segmentation.
\newblock {\em IEEE Transactions on Pattern Analysis and Machine Intelligence},
  42(5):1205--1217, 2020.

\bibitem{xu2018youtube}
Ning Xu, Linjie Yang, Yuchen Fan, Jianchao Yang, Dingcheng Yue, Yuchen Liang,
  Brian Price, Scott Cohen, and Thomas Huang.
\newblock Youtube-vos: Sequence-to-sequence video object segmentation.
\newblock In {\em ECCV}, pages 585--601, 2018.

\bibitem{yang2018efficient}
Linjie Yang, Yanran Wang, Xuehan Xiong, Jianchao Yang, and Aggelos~K
  Katsaggelos.
\newblock Efficient video object segmentation via network modulation.
\newblock In {\em CVPR}, pages 6499--6507, 2018.

\bibitem{yang2020collaborative}
Zongxin Yang, Yunchao Wei, and Yi Yang.
\newblock Collaborative video object segmentation by foreground-background
  integration.
\newblock In {\em ECCV}, 2020.

\bibitem{yang2021collaborative}
Zongxin Yang, Yunchao Wei, and Yi Yang.
\newblock Collaborative video object segmentation by multi-scale
  foreground-background integration.
\newblock {\em IEEE Transactions on Pattern Analysis and Machine Intelligence},
  2021.

\bibitem{Zeng_2019_ICCV}
Xiaohui Zeng, Renjie Liao, Li Gu, Yuwen Xiong, Sanja Fidler, and Raquel
  Urtasun.
\newblock Dmm-net: Differentiable mask-matching network for video object
  segmentation.
\newblock In {\em ICCV}, October 2019.

\bibitem{Zhang_2019_ICCV}
Lu Zhang, Zhe Lin, Jianming Zhang, Huchuan Lu, and You He.
\newblock Fast video object segmentation via dynamic targeting network.
\newblock In {\em ICCV}, October 2019.

\bibitem{zhang2020transductive}
Yizhuo Zhang, Zhirong Wu, Houwen Peng, and Stephen Lin.
\newblock A transductive approach for video object segmentation.
\newblock In {\em CVPR}, pages 6949--6958, 2020.

\bibitem{zhu2021deformable}
Xizhou Zhu, Weijie Su, Lewei Lu, Bin Li, Xiaogang Wang, and Jifeng Dai.
\newblock Deformable detr: Deformable transformers for end-to-end object
  detection.
\newblock In {\em ICLR}, 2021.

\end{thebibliography}
}
\clearpage
\includepdf[pages=1]{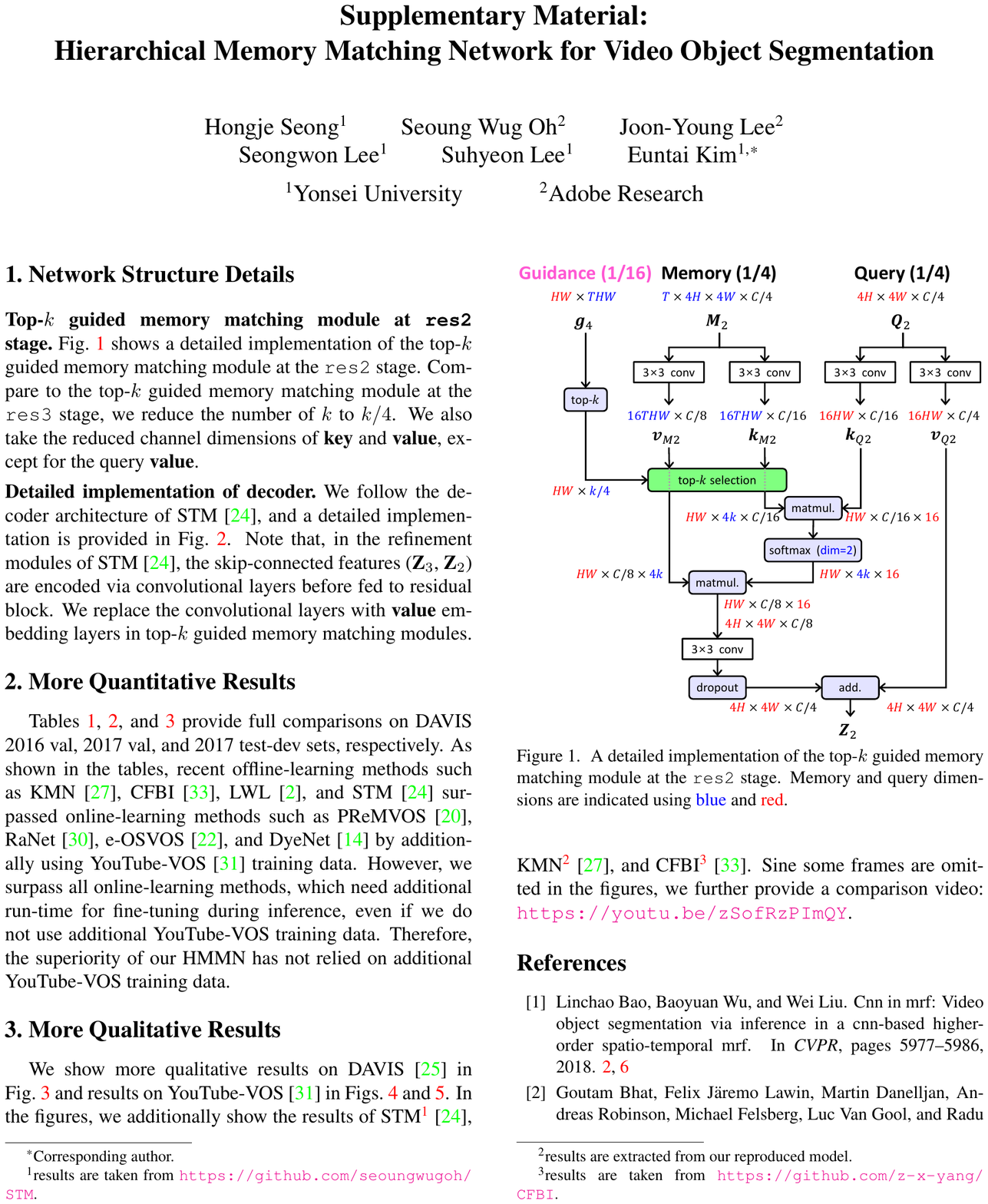}
\includepdf[pages=2]{supp.pdf}
\includepdf[pages=3]{supp.pdf}
\includepdf[pages=4]{supp.pdf}
\includepdf[pages=5]{supp.pdf}
\includepdf[pages=6]{supp.pdf}
\includepdf[pages=7]{supp.pdf}
\end{document}


\title{Supplementary Material:\\
Hierarchical Memory Matching Network for Video Object Segmentation}

\author{Hongje Seong\textsuperscript{1} \quad\quad Seoung Wug Oh\textsuperscript{2} \quad\quad Joon-Young Lee\textsuperscript{2} \\ Seongwon Lee\textsuperscript{1} \quad\quad Suhyeon Lee\textsuperscript{1} \quad\quad Euntai Kim\textsuperscript{1,}\thanks{Corresponding author.}\vspace*{0.2cm}\\
{\textsuperscript{1}Yonsei University \quad\quad\quad \textsuperscript{2}Adobe Research}}

\maketitle
\ificcvfinal\thispagestyle{empty}\fi


\vspace{-5mm}
\section{Network Structure Details}
\label{sec:1.Network_Structure_Details}
\begin{figure}[t]
\centering
\includegraphics[width=1.\linewidth]{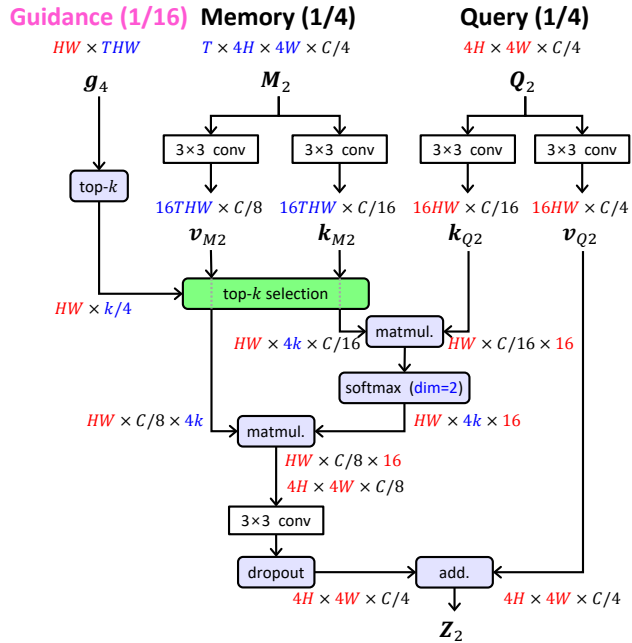}
\caption{
A detailed implementation of the top-$k$ guided memory matching module at the \texttt{res2} stage.
Memory and query dimensions are indicated using \textcolor{blue}{blue} and \textcolor{red}{red}.
}
\label{fig:topk_memory_read_detail_supp}
\end{figure}

\paragraph{Top-$k$ guided memory matching module at \texttt{res2} stage.}~Fig.~\ref{fig:topk_memory_read_detail_supp} shows a detailed implementation of the top-$k$ guided memory matching module at the \texttt{res2} stage.
Compare to the top-$k$ guided memory matching module at the \texttt{res3} stage, we reduce the number of $k$ to $k/4$.
We also take the reduced channel dimensions of \textbf{key} and \textbf{value}, except for the query \textbf{value}.

\begin{figure*}[t]
\centering
\includegraphics[width=0.8\linewidth]{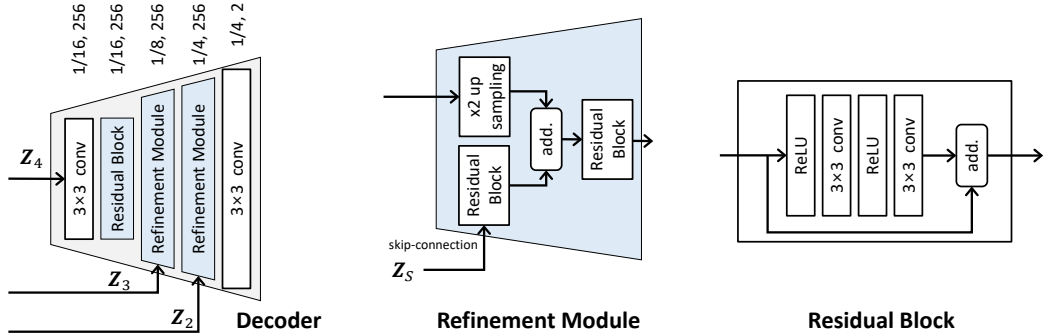}
\caption{
A detailed implementation of decoder.
We notated the output scale and channel dimension next to each block in the decoder.
}
\label{fig:decoder}
\end{figure*}

\paragraph{Detailed implementation of decoder.}~We follow the decoder architecture of STM \cite{Oh_2019_ICCV}, and a detailed implementation is provided in Fig.~\ref{fig:decoder}.
Note that, in the refinement modules of STM \cite{Oh_2019_ICCV}, the skip-connected features ($\textbf{Z}_3$, $\textbf{Z}_2$) are encoded via convolutional layers before fed to residual block.
We replace the convolutional layers with \textbf{value} embedding layers in top-$k$ guided memory matching modules.

\section{More Quantitative Results}
\label{sec:2.More_Quantitative_Results}
Tables~\ref{tab:davis2016_val_supp}, \ref{tab:davis2017_val_supp}, and \ref{tab:davis2017_test_dev_supp} provide full comparisons on DAVIS 2016 val, 2017 val, and 2017 test-dev sets, respectively.
As shown in the tables, recent offline-learning methods such as KMN~\cite{seong2020kernelized}, CFBI~\cite{yang2020collaborative}, LWL~\cite{bhat2020learning}, and STM~\cite{Oh_2019_ICCV} surpassed online-learning methods such as PReMVOS~\cite{luiten2018premvos}, RaNet~\cite{Wang_2019_ICCV}, e-OSVOS~\cite{meinhardt2020make}, and DyeNet~\cite{li2018video} by additionally using YouTube-VOS~\cite{xu2018youtube} training data.
However, we surpass all online-learning methods, which need additional run-time for fine-tuning during inference, even if we do not use additional YouTube-VOS training data.
Therefore, the superiority of our HMMN has not relied on additional YouTube-VOS training data.

\begin{table}[t]
\begin{center}
\footnotesize
\begin{tabular}{lccccc}
\toprule
Method                                                  & OL         & $\mathcal{J\&F}$      & $\mathcal{J}$         & $\mathcal{F}$         & Time     \\
\midrule
OSVOS \cite{caelles2017one}                             & \checkmark & 80.2                  & 79.8                  & 80.6                  & 9$s$     \\
MaskRNN \cite{hu2017maskrnn}                            & \checkmark & 80.8                  & 80.7                  & 80.9                  & -        \\
VidMatch   \cite{hu2018videomatch}                      &            & -                     & 81.0                  & -                     & 0.32$s$  \\
FAVOS \cite{cheng2018fast}                              &            & 81.0                  & 82.4                  & 79.5                  & 1.8$s$   \\
LSE \cite{ci2018video}                                  & \checkmark & 81.6                  & 82.9                  & 80.3                  & -        \\
FEELVOS   \cite{voigtlaender2019feelvos}                &            & 81.7                  & 80.3                  & 83.1                  & 0.45$s$  \\
FEELVOS \textbf{(+YV)}   \cite{voigtlaender2019feelvos} &            & 81.7                  & 81.1                  & 82.2                  & 0.45$s$  \\
FRTM   \cite{robinson2020learning}                      & \checkmark           & 81.7                  & -                     & -                     & 0.05$s$  \\
RGMP \cite{wug2018fast}                                 &            & 81.8                  & 81.5                  & 82.0                  & 0.13$s$  \\
A-GAME \textbf{(+YV)}   \cite{johnander2019generative}  &            & -                     & 82.0                  & -                     & 0.07$s$  \\
SAT \cite{chen2020state}                                &            & 83.1                  & 82.6                  & 83.6                  & 0.03$s$  \\
FRTM \textbf{(+YV)}   \cite{robinson2020learning}       & \checkmark           & 83.5                  & -                     & -                     & 0.05$s$  \\
DTN \cite{Zhang_2019_ICCV}                              &            & 83.6                  & 83.7                  & 83.5                  & 0.07$s$  \\
CINN \cite{bao2018cnn}                                  & \checkmark & 84.2                  & 83.4                  & 85.0                  & $>$30$s$ \\
DyeNet \cite{li2018video}                               &            & -                     & 84.7                  & -                     & 0.42$s$  \\
RaNet \cite{Wang_2019_ICCV}                             &            & 85.5                  & 85.5                  & 85.4                  & 0.03$s$  \\
OnAVOS   \cite{voigtlaender2017online}                  & \checkmark & 85.5                  & 86.1                  & 84.9                  & 13$s$    \\
STG-Net   \cite{liu2020spatiotemporal}                  &            & 85.7                  & 85.4                  & 86.0                  & 0.16$s$  \\
OSVOS$^S$   \cite{maninis2018video}                     & \checkmark & 86.0                  & 85.6                  & 86.4                  & 4.5$s$   \\
DIPNet \cite{hu2020dipnet}                              & \checkmark & 86.1                  & 85.8                  & 86.4                  & 1.09$s$  \\
CFBI   \cite{yang2020collaborative}                     &            & 86.1                  & 85.3                  & 86.9                  & 0.18$s$  \\
STM \cite{Oh_2019_ICCV}                                 &            & 86.5                  & 84.8                  & 88.1                  & 0.16$s$  \\
PReMVOS   \cite{luiten2018premvos}                      & \checkmark & 86.8                  & 84.9                  & 88.6                  & 32.8$s$  \\
e-OSVOS   \cite{meinhardt2020make}                      & \checkmark & 86.8                  & 86.6                  & 87.0                  & 3.4$s$   \\
DyeNet \cite{li2018video}                               & \checkmark & -                     & 86.2                  & -                     & 2.32$s$  \\
RaNet \cite{Wang_2019_ICCV}                             & \checkmark & 87.1                  & 86.6                  & 87.6                  & 4$s$     \\
KMN \cite{seong2020kernelized}                          &            & 87.6                  & 87.1                  & 88.1                  & 0.12$s$  \\
STM \textbf{(+YV)}   \cite{Oh_2019_ICCV}                &            & 89.3                  & 88.7                  & 89.9                  & 0.16$s$  \\
CFBI \textbf{(+YV)}   \cite{yang2020collaborative}      &            & 89.4                  & 88.3                  & 90.5                  & 0.18$s$  \\
KMN \textbf{(+YV)}   \cite{seong2020kernelized}         &            & 90.5                  & 89.5                  & 91.5                  & 0.12$s$  \\
\midrule
HMMN                                                    &            & 89.4 & 88.2 & 90.6  & 0.10$s$  \\
HMMN \textbf{(+YV)}                                     &            & \textbf{90.8}         & \textbf{89.6}         & \textbf{92.0}         & 0.10$s$ \\
\bottomrule
\end{tabular}
\end{center}
\caption{Full comparison on DAVIS 2016 validation set.
(\textbf{+YV}) indicates YouTube-VOS is additionally used for training, and OL denotes the use of online-learning strategies during test-time. Time measurements reported in this table are directly from the corresponding papers. 
}
\label{tab:davis2016_val_supp}
\end{table}

\begin{table}[t]
\begin{center}
\footnotesize
\begin{tabular}{lcccc}
\toprule
Method                                                  & OL         & $\mathcal{J\&F}$ & $\mathcal{J}$ & $\mathcal{F}$ \\
\midrule
OSVOS \cite{caelles2017one}                             & \checkmark & 60.3             & 56.6          & 63.9          \\
VidMatch   \cite{hu2018videomatch}                      &            & 62.4             & 56.5          & 68.2          \\
MaskRNN \cite{hu2017maskrnn}                            & \checkmark & -                & 60.5          & -             \\
RaNet \cite{Wang_2019_ICCV}                             &            & 65.7             & 63.2          & 68.2          \\
AGSS-VOS \cite{Lin_2019_ICCV}                           &            & 66.6             & 63.4          & 69.8          \\
RGMP \cite{wug2018fast}                                 &            & 66.7             & 64.8          & 68.6          \\
DTN \cite{Zhang_2019_ICCV}                              &            & 67.4             & 64.2          & 70.6          \\
AGSS-VOS \textbf{(+YV)}   \cite{Lin_2019_ICCV}          &            & 67.4             & 64.9          & 69.9          \\
OnAVOS   \cite{voigtlaender2017online}                  & \checkmark & 67.9             & 64.5          & 71.2          \\
OSVOS$^S$   \cite{maninis2018video}                     & \checkmark & 68.0             & 64.7          & 71.3          \\
DIPNet \cite{hu2020dipnet}                              & \checkmark & 68.5             & 65.3          & 71.6          \\
FRTM   \cite{robinson2020learning}                      & \checkmark           & 68.8             & -             & -             \\
FEELVOS   \cite{voigtlaender2019feelvos}                &            & 69.1             & 65.9          & 72.3          \\
DyeNet \cite{li2018video}                               &            & 69.1             & 67.3          & 71.0          \\
A-GAME \textbf{(+YV)}   \cite{johnander2019generative}  &            & 70.0             & 67.2          & 72.7          \\
CINN \cite{bao2018cnn}                                  & \checkmark & 70.7             & 67.2          & 74.2          \\
DMM-Net \cite{Zeng_2019_ICCV}                           &            & 70.7             & 68.1          & 73.3          \\
GC \cite{li2020fast}                                    &            & 71.4             & 69.3          & 73.5          \\
STM \cite{Oh_2019_ICCV}                                 &            & 71.6             & 69.2          & 74.0          \\
FEELVOS \textbf{(+YV)}   \cite{voigtlaender2019feelvos} &            & 72.0             & 69.1          & 74.0          \\
SAT \cite{chen2020state}                                &            & 72.3             & 68.6          & 76.0          \\
TVOS   \cite{zhang2020transductive}                     &            & 72.3             & 69.9          & 74.7          \\
LWL \cite{bhat2020learning}                             &            & 74.3             & 72.2          & 76.3          \\
AFB+URR \cite{liang2020video}                           &            & 74.6             & 73.0          & 76.1          \\
STG-Net   \cite{liu2020spatiotemporal}                  &            & 74.7             & 71.5          & 77.9          \\
CFBI   \cite{yang2020collaborative}                     &            & 74.9             & 72.1          & 77.7          \\
DTTM-TAN \cite{huang2020fast}                           &            & 75.9             & 72.3          & 79.4          \\
KMN \cite{seong2020kernelized}                          &            & 76.0             & 74.2          & 77.8          \\
FRTM \textbf{(+YV)}   \cite{robinson2020learning}       & \checkmark           & 76.7             & -             & -             \\
e-OSVOS   \cite{meinhardt2020make}                      & \checkmark & 77.2             & 74.4          & 80.0          \\
PReMVOS   \cite{luiten2018premvos}                      & \checkmark & 77.8             & 73.9          & 81.7          \\
LWL \textbf{(+YV)}   \cite{bhat2020learning}            &            & 81.6             & 79.1          & 84.1          \\
STM \textbf{(+YV)}   \cite{Oh_2019_ICCV}                &            & 81.8             & 79.2          & 84.3          \\
CFBI \textbf{(+YV)}   \cite{yang2020collaborative}      &            & 81.9             & 79.1          & 84.6          \\
EGMN \textbf{(+YV)}   \cite{lu2020video}                &            & 82.8             & 80.2          & 85.2          \\
KMN \textbf{(+YV)}   \cite{seong2020kernelized}         &            & 82.8             & 80.0          & 85.6          \\
\midrule
HMMN                                                    &            & 80.4             & 77.7          & 83.1          \\
HMMN \textbf{(+YV)}                                     &            & \textbf{84.7}    & \textbf{81.9} & \textbf{87.5}\\
\bottomrule
\end{tabular}
\end{center}
\caption{Full comparison on DAVIS 2017 validation set.
}
\label{tab:davis2017_val_supp}
\end{table}

\begin{table}[t]
\begin{center}
\footnotesize
\centering
\begin{tabular}{lcccc}
\toprule
Method                                                  & OL         & $\mathcal{J\&F}$ & $\mathcal{J}$ & $\mathcal{F}$ \\
\midrule
OSMN \cite{yang2018efficient}                           &            & 39.3             & 33.7          & 44.9          \\
FAVOS \cite{cheng2018fast}                              &            & 43.6             & 42.9          & 44.2          \\
OSVOS \cite{caelles2017one}                             & \checkmark & 50.9             & 47.0          & 54.8          \\
CapsuleVOS   \cite{Duarte_2019_ICCV}                    &            & 51.3             & 47.4          & 55.2          \\
OnAVOS   \cite{voigtlaender2017online}                  & \checkmark & 52.8             & 49.9          & 55.7          \\
RGMP \cite{wug2018fast}                                 &            & 52.9             & 51.3          & 54.4          \\
RaNet \cite{Wang_2019_ICCV}                             &            & 53.4             & 55.3          & 57.2          \\
OSVOS$^S$   \cite{maninis2018video}                     & \checkmark & 57.5             & 52.9          & 62.1          \\
FEELVOS \textbf{(+YV)}   \cite{voigtlaender2019feelvos} &            & 57.8             & 55.1          & 60.4          \\
TVOS   \cite{zhang2020transductive}                     &            & 63.1             & 58.8          & 67.4          \\
STG-Net   \cite{liu2020spatiotemporal}                  &            & 63.1             & 59.7          & 66.5          \\
e-OSVOS   \cite{meinhardt2020make}                      & \checkmark & 64.8             & 60.9          & 68.6          \\
DTTM-TAN \cite{huang2020fast}                           &            & 65.4             & 61.3          & 70.3          \\
Lucid \cite{khoreva2019lucid}                           & \checkmark & 66.7             & 63.4          & 69.9          \\
CINN \cite{bao2018cnn}                                  & \checkmark & 67.5             & 64.5          & 70.5          \\
DyeNet \cite{li2018video}                               & \checkmark & 68.2             & 65.8          & 70.5          \\
PReMVOS   \cite{luiten2018premvos}                      & \checkmark & 71.6             & 67.5          & 75.7          \\
STM \textbf{(+YV)}   \cite{Oh_2019_ICCV}                &            & 72.2             & 69.3          & 75.2          \\
CFBI \textbf{(+YV)}   \cite{yang2020collaborative}      &            & 74.8             & 71.1          & 78.5          \\
KMN \textbf{(+YV)}   \cite{seong2020kernelized}         &            & 77.2             & 74.1          & 80.3          \\
\midrule
HMMN \textbf{(+YV)}                                     &            & \textbf{78.6}    & \textbf{74.7} & \textbf{82.5}\\
\bottomrule
\end{tabular}
\end{center}
\caption{Full comparison on DAVIS 2017 test-dev set.
}
\label{tab:davis2017_test_dev_supp}
\end{table}

\section{More Qualitative Results}
\label{sec:3.More_Qualitative_Results}
\begin{figure*}[t]
\centering
\includegraphics[width=1\linewidth]{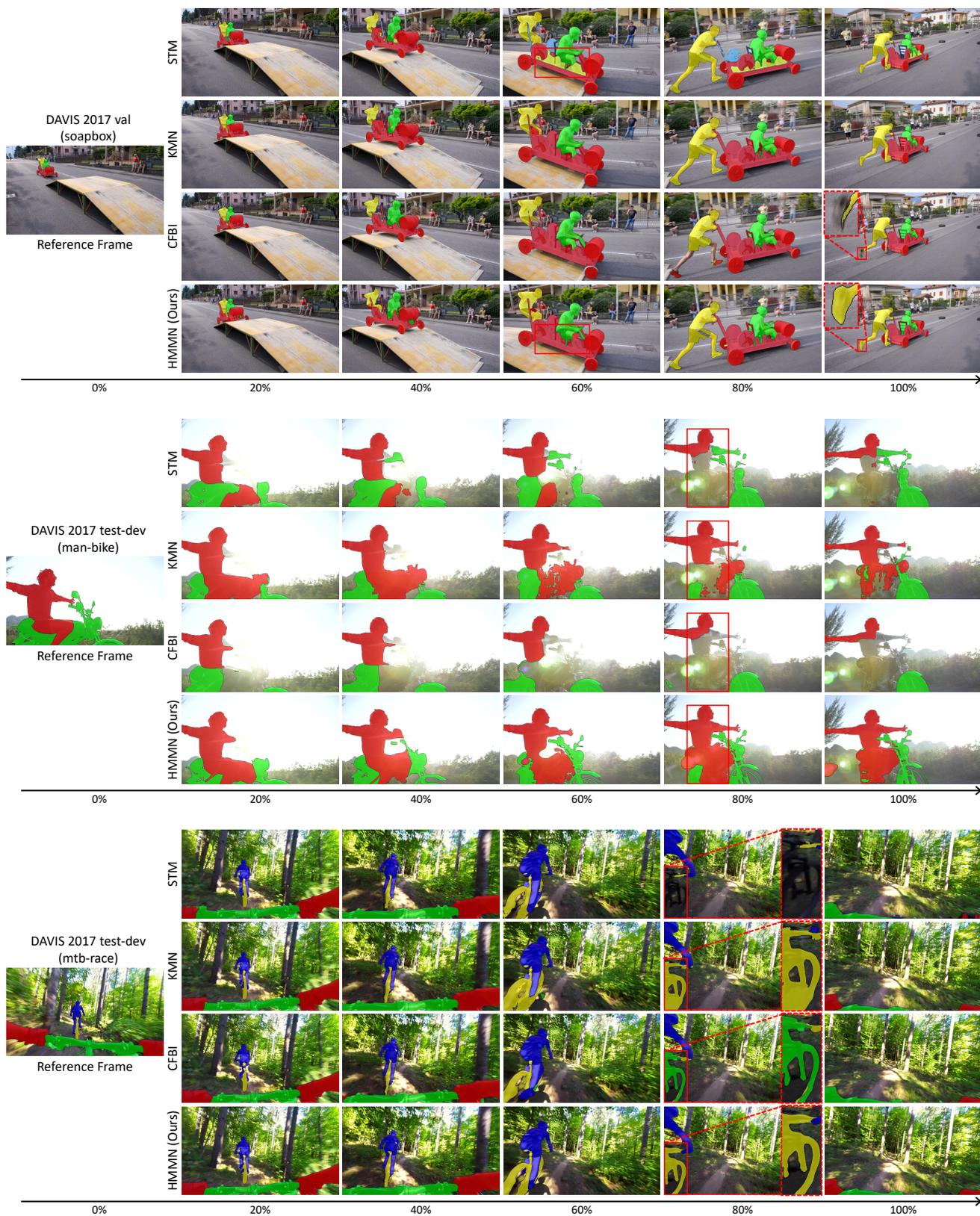}
\caption{More qualitative results on DAVIS 2017 validation and test-dev sets.
We marked significant improvements from STM \cite{Oh_2019_ICCV}, KMN \cite{seong2020kernelized}, and CFBI \cite{yang2020collaborative} using red boxes.
\vspace{-1cm}
}
\label{fig:qualitative_results_1}
\end{figure*}

We show more qualitative results on DAVIS~\cite{pont20172017} in Fig.~\ref{fig:qualitative_results_1} and results on YouTube-VOS~\cite{xu2018youtube} in Figs.~\ref{fig:qualitative_results_2} and~\ref{fig:qualitative_results_3}.
In the figures, we additionally show the results of STM\footnote{results are taken from \url{https://github.com/seoungwugoh/STM}.}~\cite{Oh_2019_ICCV}, KMN\footnote{results are extracted from our reproduced model.}~\cite{seong2020kernelized}, and CFBI\footnote{results are taken from \url{https://github.com/z-x-yang/CFBI}.}~\cite{yang2020collaborative}.
Sine some frames are omitted in the figures, we further provide a comparison video:  \url{https://youtu.be/zSofRzPImQY}.

\begin{figure*}[t]
\centering
\includegraphics[width=1\linewidth]{figures/qualitative_results_supp_2.pdf}
\caption{More qualitative results on YouTube-VOS 2019 validation set.
}
\label{fig:qualitative_results_2}
\end{figure*}

\begin{figure*}[t]
\centering
\includegraphics[width=1\linewidth]{figures/qualitative_results_supp_3.pdf}
\caption{More qualitative results on YouTube-VOS 2019 validation set.
}
\label{fig:qualitative_results_3}
\end{figure*}

{\small
\bibliographystyle{ieee_fullname}
\bibliography{main-supp}
}